\documentclass{IEEEtran}
\usepackage{cite}
\usepackage{amsmath,amssymb,amsfonts}
\usepackage{graphicx}
\usepackage{textcomp}

\usepackage{booktabs}
\usepackage{multirow}
\usepackage{mathtools}

\usepackage{amsmath,amsfonts,bm}

\def\eqref#1{equation~\ref{#1}}

\def\1{\bm{1}}

\def\vh{{\bm{h}}}

\def\vx{{\bm{x}}}

\def\vz{{\bm{z}}}

\DeclareMathAlphabet{\mathsfit}{\encodingdefault}{\sfdefault}{m}{sl}
\SetMathAlphabet{\mathsfit}{bold}{\encodingdefault}{\sfdefault}{bx}{n}

\usepackage{siunitx}
\usepackage{algorithm}
\usepackage{algpseudocode}
\usepackage{nicefrac}

\usepackage{pifont}
\newcommand{\xmark}{\ding{55}}  

\usepackage{tikz}

\usepackage{color}
\usepackage[normalem]{ulem}

\def\BibTeX{{\rm B\kern-.05em{\sc i\kern-.025em b}\kern-.08em
    T\kern-.1667em\lower.7ex\hbox{E}\kern-.125emX}}
\begin{document}
\title{Stochastic Siamese MAE Pretraining for Longitudinal Medical Images}
\author{Taha Emre, Arunava Chakravarty, Thomas Pinetz, Dmitrii Lachinov, Martin J. Menten, Hendrik Scholl, Sobha Sivaprasad, Daniel Rueckert, Andrew Lotery, Stefan Sacu, Ursula Schmidt-Erfurth, Hrvoje Bogunovi\'c, for the PINNACLE Consortium, for the Alzheimer’s Disease Neuroimaging Initiative (ADNI)
\thanks{This work was supported in part by Wellcome Trust Collaborative Award (PINNACLE) Ref. 210572/Z/18/Z, and has received support from the European Union, EIC-2023-PATHFINDEROPEN-01 (I-SCREEN, grant no. 101130093). Funded by the European Union. Views and opinions expressed are however those of
the author(s) only and do not necessarily reflect those of the European Union or
European Innovation Council and SMEs Executive Agency (EISMEA). Neither the
European Union nor the granting authority can be held responsible for them. For the purpose of open access, the author has applied a CC-BY public copyright licence to any author accepted manuscript version arising from this submission.}  
\thanks{Taha Emre, Thomas Pinetz, Dmitrii Lachinov, and Hrvoje Bogunovi\'{c} are with the Institute of Artificial Intelligence, Center for Medical Data Science, Medical University of Vienna, Austria (taha.emre@meduniwien.ac.at, hrvoje.bogunovic@meduniwien.ac.at).} 
\thanks{Arunava Chakravarty and Stefan Sacu are with the Department of Ophthalmology and Optometry, Medical University of Vienna, Austria.}
\thanks{Ursula Schmidt-Erfurth is  with Ophthalmic Image Analysis Group (OPTIMA), Medical University of Vienna, Austria.}
\thanks{Martin J. Menten and Daniel Rueckert are with BioMedIA, Department of Computing, Imperial College London, London, United Kingdom, and 
Chair for AI in Healthcare and Medicine, Technical University of Munich, Munich, Germany.}
\thanks{Sobha Sivaprasad is with Moorfields National Institute for Health and Care Biomedical Research Centre, Moorfields Eye Hospital, London, United Kingdom, and Institute of Ophthalmology, University College London, London, United Kingdom.}
\thanks{Hendrik P.N. Scholl is with Department of Clinical Pharmacology, Medical University of Vienna, Vienna, Austria, and Pallas Kliniken AG, Pallas Klinik Zürich, Zürich, Switzerland, and European Vision Institute, Basel, Basel-Stadt, Switzerland.}
\thanks{Andrew Lotery is with Faculty of Medicine, University of Southampton, Southampton, Hampshire, United Kingdom.}
\thanks{Data used in preparation of this article were obtained from the Alzheimer’s Disease Neuroimaging Initiative (ADNI) database (adni.loni.usc.edu). As such, the investigators within the ADNI  contributed to the design and implementation of ADNI and/or provided data but did not participate in analysis or writing of this report. A complete listing  of ADNI investigators can be found at: http://adni.loni.usc.edu/wp-content/uploads/how\_to\_apply/ADNI\_Acknowledgement\_List.pdf}
}

\maketitle

\begin{abstract}

Temporally aware image representations are crucial for capturing disease progression in 3D volumes of longitudinal medical datasets. However, recent state‑of‑the‑art self‑supervised learning approaches like Masked Autoencoding (MAE), despite their strong representation learning capabilities, lack temporal awareness. In this paper, we propose STAMP (\textbf{S}tochastic \textbf{T}emporal \textbf{A}utoencoder with \textbf{M}asked \textbf{P}retraining), a Siamese MAE framework that encodes temporal information through a stochastic process by conditioning on the time difference between the 2 input volumes. Unlike deterministic Siamese approaches, which compare scans from different time points but fail to account for the inherent uncertainty in disease evolution, STAMP learns temporal dynamics stochastically by reframing the MAE reconstruction loss as a conditional variational inference objective. We evaluated STAMP on two OCT and one MRI datasets with multiple visits per patient. STAMP pretrained ViT models outperformed both existing temporal MAE methods and foundation models on different late stage Age‑Related Macular Degeneration and Alzheimer’s Disease progression prediction which require models to learn the underlying non‑deterministic temporal dynamics of the diseases.

\end{abstract}

\begin{IEEEkeywords}
Masked-Autoencoding, Variational-Inference, Self-supervised-learning, Disease Progression.
\end{IEEEkeywords}

\section{Introduction}
\label{sec:introduction}

Longitudinal medical images are routinely acquired in hospitals across multiple visits for each patient to track degenerative disease progression and inform clinical decisions. In ophthalmology, Age-related macular degeneration (AMD), a major contributor to vision loss and blindness~\cite{bressAMD}, causes irreversible tissue damage with the severity steadily worsening over time, and is diagnosed using 3D optical coherence tomography (OCT) scans. Intermediate stage AMD (iAMD) may follow one of two pathways: neovascular AMD (wet-AMD) and Geographic Atrophy (GA) (Fig.~\ref{disease} - 1st \& 2nd rows), indicating multiple prognostic paths~\cite{hallak2019imaging}. Moreover, progression to late stage wet-AMD can occur suddenly (Fig.~\ref{disease} - 1st row), contributing to the uncertainty. Similarly, Alzheimer’s disease (AD) is an irreversible, progressive neurodegenerative disorder and the leading cause of dementia. With approximately 50 million people affected worldwide~\cite{breijyeh2020comprehensive}, early and accurate detection is critical. While current treatments only address symptoms, volumetric brain MRI is widely used in clinical settings to identify early structural changes in patients with cognitively normal (CN) and mild cognitive impairment (MCI) diagnoses to make informed clinical decisions and understand the onset of dementia~\cite{bucholc2023hybrid} (Fig.~\ref{disease} - 3rd row).


The ability to forecast the future trajectory of disease progression from the current patient scan is crucial for personalized treatment~\cite{BERG2015146}. It can help prioritize patients at increased risk of converting to the advanced stage through timely treatment and frequent monitoring~\cite{yim2020predicting, ansart2021predicting}. However, manual identification of high-risk patients is often subjective, tedious, and error-prone. Moreover, the high cost of diagnostic labeling often leaves longitudinal data unlabeled, limiting their use for forecasting. 
This underscores the need to develop automated methods that can utilize unlabeled longitudinal medical datasets.

Self-supervised learning (SSL) aims to leverage available unlabeled data to learn meaningful features about disease. In the longitudinal setting, SSL methods learn to model temporal relationships by comparing images from different time points without relying on any human supervision. Although most existing longitudinal SSL methods make use of multiple visits as input for future forecasting, they ultimately learn to extract deterministic temporal features. However, this is limiting because the rate of disease progression and the resulting anatomical changes vary significantly across patients. Therefore, it becomes crucial to model multiple possible future outcomes as a stochastic variable, allowing models to predict a distribution over a range of plausible disease progression trajectories. Leveraging large longitudinal medical datasets,  we can model these temporal dynamics in an unsupervised manner so that future progression can be estimated from a single visit, reducing hospital load and enabling timely treatment planning.

In this work, we propose STAMP (Stochastic Temporal Autoencoder with Masked Pretraining), a novel SSL approach that extends masked auto-encoding to longitudinal medical imaging data within a Siamese framework. We hypothesize that latent factors driving temporal evolution can be inferred without explicit supervision by contrasting a patient's two visits via a variational process. To account for the inherent uncertainty arising from these latent factors, we sample temporal feature tokens from a learned prior probability distribution conditioned on the time interval and the current input scan. STAMP is therefore able to capture long-term non-deterministic temporal dependencies using only two visits. Moreover, our SSL does not require spatio-temporal attention to learn temporal dynamics~\cite{bertasius2021space}, making it model agnostic and compatible with standard Vision Transformer backbones. Our key contributions are\footnote{Source code is available at https://github.com/EmreTaha/STAMP}:



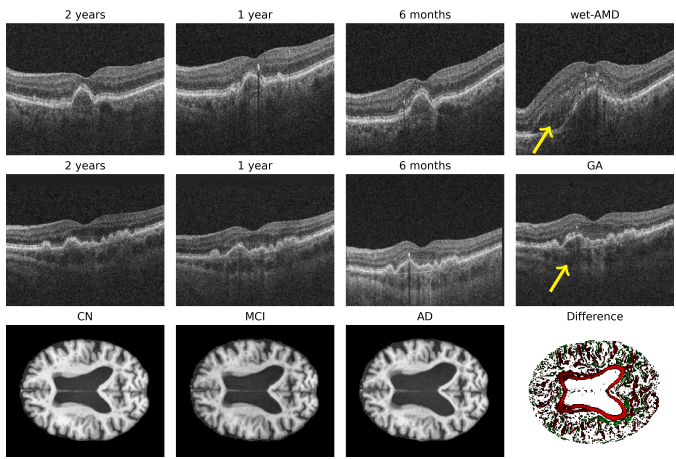
\begin{figure}[!t]
  \centering
  \begin{tikzpicture}
    \node[anchor=south west, inner sep=0] (img)
      at (0,0) {\includegraphics[width=\linewidth]{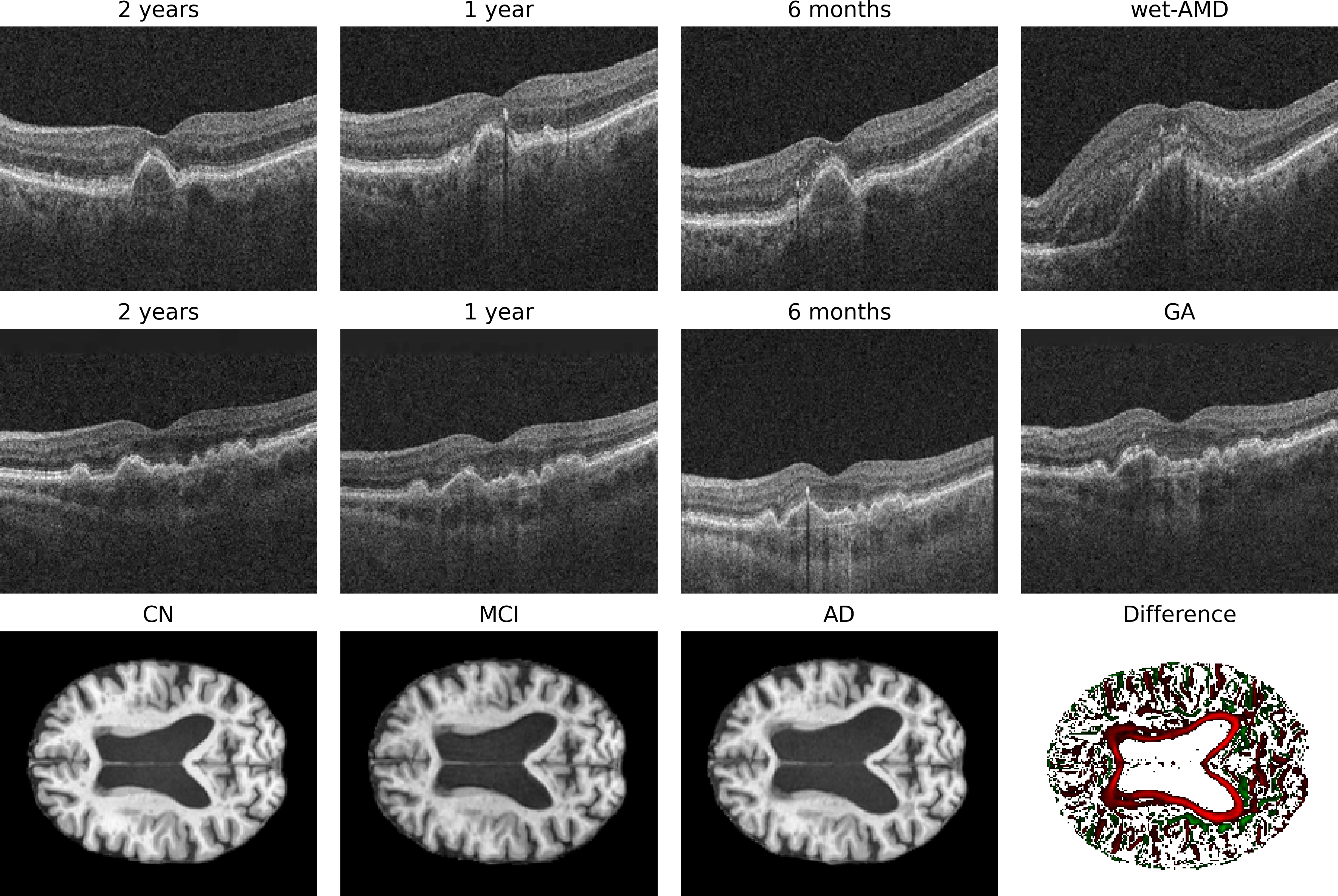}};
    \begin{scope}[x={(img.south east)}, y={(img.north west)}]
      \foreach \c/\xA/\yA/\xB/\yB in {
        4/0.79/0.34+0.34/0.815/0.41+0.33    
      } {
        \draw[->, line width=1.2pt, yellow]
             (\xA,\yA) -- (\xB,\yB);
      }

      \foreach \c/\xA/\yA/\xB/\yB in {
        4/0.815/0.05+0.32/0.84/0.12+0.31    
      } {
        \draw[->, line width=1.2pt, yellow]
             (\xA,\yA) -- (\xB,\yB);
      }
    \end{scope}
  \end{tikzpicture}
\caption{\textbf{Top}: HARBOR dataset for wet-AMD conversion from iAMD, arrow pointing the subretinal fluid. \textbf{Middle}: PINNACLE dataset for GA conversion from iAMD, arrow pointing increasing light transmission due to atrophy. \textbf{Bottom}: ADNI dataset, CN to MCI to AD with difference map, red/green means decrease/increase in pixel intensity.} \label{disease}
\end{figure}

\begin{itemize}
\item \textbf{Time Awareness:} We introduce a learnable encoding of time difference between input images, improving temporal image features over longer intervals.



\item \textbf{Time-Conditioned Stochasticity:} We introduce an inference-time stochastic formulation with a time-conditioned prior, learned from the current scan and prediction horizon, and a masked-future posterior used during pretraining. This enables stochastic future representations to be sampled from a single scan at inference.

\item \textbf{3D Volumetric Extension:} We employ a fully 3D approach to learn representations of volumetric scans at each time point, marking the first adaptation of the Siamese MAE framework to 3D medical scans.
\item \textbf{Extensive Evaluation:} We evaluate STAMP in two different modalities (OCT and MRI) on various prognostic tasks with different predictive time windows.
\end{itemize}

\section{Background And Related work}

Self-supervised learning is a pretraining approach that extracts meaningful representations from data without manual labels, thereby improving downstream performance when annotations are scarce. In medical imaging, SSL has been used for various downstream tasks, such as segmentation, disease classification, or disease forecasting~\cite{huang2023self}. Longitudinal medical datasets offer a unique opportunity to learn long-term relations with SSL.

\subsubsection*{Longitudinal SSL}
Previously, various longitudinal SSL methods have been developed using contrastive learning (temporal sensitivity~\cite{emre2024learning}, age as metadata~\cite{dufumier2021contrastive}), and  pretext learning (time difference prediction~\cite{rivail2019modeling},  patient-level trajectory modeling~\cite{ren2022local}). Although these models achieved considerable results, they are limited by their reliance on metadata and by the large batch-size dependence of contrastive learning~\cite{chen2020simple}, which makes scaling to longitudinal 3D volumes computationally inefficient. 

Recent temporal medical imaging models further demonstrate the importance of longitudinal information. Holste et al.~\cite{holste2024harnessing} proposed supervised Transformer-based sequence modeling for eye-disease prognosis using longitudinal image sequences, showing the benefit of explicitly modeling prior visits. Huo et al.~\cite{huo2025time} proposed time-to-event pretraining for 3D medical imaging by leveraging additional clinical information such as EHR-derived events and text reports. In contrast, STAMP focuses on image-only self-supervised pretraining from longitudinal 3D image pairs. It does not require event labels, reports, or multi-visit inputs at inference, and instead supports single-scan, time-prompted prognosis through a learned time-conditioned stochastic prior.

\subsubsection*{Masked Auto-Encoding}
However, with the advent of Vision Transformer (ViT) architectures, Masked Auto-encoding (MAE)~\cite{he2022masked} has emerged as the state-of-the-art SSL method for pretraining on unlabeled data. It addresses ViTs’ lack of inductive biases such as spatial locality and translation equivariance by adapting an image inpainting task~\cite{pathak2016context} to the patch‑based attention mechanism of ViTs. After dividing each input image into non‑overlapping patches (\textit{tokens}), MAE randomly masks up to 90\% of them, and only the visible patches are processed by a ViT encoder to extract strong high‑ and low‑level feature embeddings~\cite{kong2023understanding}. A lightweight ViT‑based decoder then reconstructs the masked patches from the visible patch embeddings combined with learnable mask tokens and is trained with an $L_2$ loss. MAE has become the standard SSL pretraining step for a variety of domains, including medical imaging foundation models~\cite{zhou2023retfound,visionfm}. Yet vanilla MAE‑pretrained models lack the critical temporal information required for video or disease progression‑based downstream tasks.

\subsubsection*{MAE-based Longitudinal SSL}
Longitudinal MAE has also been proposed for video datasets~\cite{feichtenhofer2022masked,tong2022videomae} by stacking and masking frames to exploit temporal continuity and learn the temporal information between them. The longitudinal frames are treated as a volumetric input, and masking is done along both spatial and temporal dimensions.
 DropMAE~\cite{wu2023dropmae} aims to improve upon \cite{feichtenhofer2022masked} by dropping tokens from different frames that have high attention scores. This forces the network to use tokens from different frames, improving temporal reasoning. In comparison, longitudinal medical imaging data acquired across multiple patient visits presents unique challenges for SSL in general, and MAE in particular due to the temporal aspect involved in disease progression~\cite{YANG20191}. As a solution, these methods were adapted to medical images by stacking 2D scans from different time-points~\cite{zhang2024whole,zeghlache2025mae} with longitudinal input specific model architectures. While these approaches improve temporal tasks, they do not utilize cross-time completion to learn temporal relations, instead relying on stacking temporal images or frames to 3D inputs, which increases computation substantially and limits the inference to time-series input. In particular, stacking 3D medical volumes into 4D inputs further increases computational cost significantly.

Siamese Masked Autoencoders (SiamMAE)~\cite{gupta2023siamese} improve temporal features using cross-time frame completion. It comprises two ViT encoders with shared weights (Siamese), and a cross-attention based decoder, reducing computational complexity. Two randomly sampled frames from a video are patchified and the future frame is masked similar to MAE. The encoder processes the unmasked past frame and the visible patches of the future frame. A learnable mask token is attached to each masked patch position in the future frame. The decoder performs cross-time completion by \textit{querying} the visible patch embeddings and mask tokens, against the fully visible patch embeddings of the past frame, which provide the $Key$ and $Value$ for cross-attention. This allows the network to reconstruct the future by comparing a fully visible past with a partially visible future, while improving computational efficiency compared to self-attention. After pretraining, the encoder is evaluated by feeding a single frame to predict the label of a future frame. Siamese MAE architecture is also used for temporal learning by mimicking the temporal changes via augmentation and reconstructing the augmented image~\cite{eymael2025efficient} without temporal inputs. Similar Siamese MAE-based SSL has been used for 3D scene generation from 2D images of different angles~\cite{weinzaepfel2022croco}, multi-scale representation learning~\cite{tang2023crossscale} and camera depth estimation~\cite{wang2024dust3r}. Although these methods leverage cross-time completion to learn temporal dynamics, their determinism limits them to modeling the mean behavior, preventing learning of multiple outcome scenarios.


\subsubsection*{Stochastic Representation Learning}
While the deterministic SiamMAE captures temporal information efficiently, its deterministic nature fails when the future is fundamentally uncertain. In case of abrupt changes or progression, such as ball bouncing off a wall or the rapid AMD conversion to a late disease stage, these models tend to embed the average of possible outcomes, highlighting the need for stochasticity. As a pioneering work, Denton et al.~\cite{denton2018stochastic} proposed to learn a stochastic prior from past frame embeddings to tackle the multi-outcome issue, removing the reliance on a fixed distribution used in Variational Autoencoders (VAE)~\cite{kingma2013auto}. The KL-divergence is computed between the learned prior and the posterior (derived from the future frame). Later on, it is adapted to ViT using stochastic tokens for action generation~\cite{petrovich2021action} by learning a stochastic prior token alongside the previous action embeddings in the attention blocks. Recently, RSP~\cite{jang2024visual} introduced a learnable prior in SiamMAE pretraining for videos. The authors additionally proposed to learn a target distribution, denoted as posterior, which is derived from the  past and future frames without masking to ensure a meaningful posterior distribution. The prior, which is only based on the past, is then aligned by penalizing the KL divergence between the two distributions~\cite{denton2018stochastic}, similar to the common practice in reinforcement learning~\cite{hafner2023mastering, schmidt2024learning, micheli2024efficient}. Without masking, the model lacked the spatial knowledge, requiring another additional masked encoder pass to compensate, increasing the computational cost substantially. 

The aforementioned models incorporate stochasticity via time series input or multiple encoder passes, increasing computation and ambiguity in predictions as the time gaps grow. Especially RSP suffers from matching a time-independent prior to a time-dependent posterior which leads to time-insensitivity at inference and limits application to only close temporal proximity. As a result, these methods perform suboptimally on longitudinal medical data since intervals between patient visits can span several months or even years rather than seconds. This leads to two key differences between longitudinal medical images and natural videos: \textbf{(i)} longer intervals may require to explicitly encode the temporal information during representation learning; \textbf{(ii)} wide inter-subject variability in disease progression rates due to unknown latent factors (e.g. genetics, dietary habits) makes reconstruction of a single deterministic future state insufficient. STAMP addresses these points by reformulating SiamMAE as a time conditional variational inference. It directly learns a stochastic latent space conditioned on both the time interval and past embeddings, eliminating additional reconstruction tasks or time unaware stochasticity as proposed in RSP. Crucially, this formulation allows both stochasticity and time-awareness to be exploited at inference for downstream tasks.


\begin{figure*}[tbh]
\centering
\includegraphics[width=0.95\linewidth]{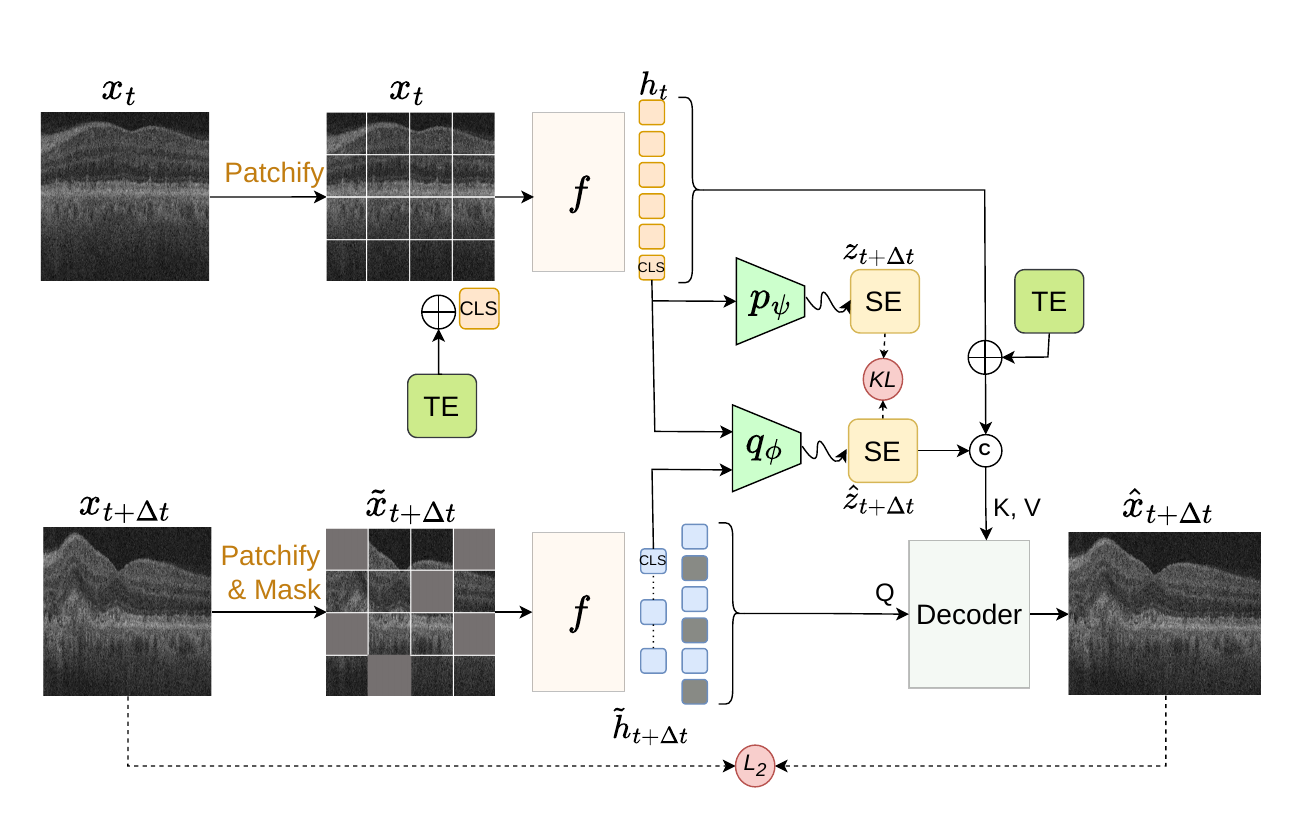}
\caption{An overview of STAMP. Two 3D volumes ($\vx_t$ and $\vx_{t+\Delta t}$) $\Delta t$ apart from a patient are used as input. After patchifying both scans, only the future visit is masked ($\Tilde{\vx}_{t+\Delta t}$) and a learnable CLS token is attached to both branches. \textit{TE} and the subsequent summation indicate Temporal Encoding (\textit{TE}) added to the CLS token with $\Delta t$ prompting in $x_t$ branch. A ViT-based encoder $f$ embeds the past visit and visible patches of the future into $h_t$ and $\Tilde{h}_{t+\Delta t}$. For the stochasticity, the posterior ($q_\phi$) is learned from the embeddings $CLS_t$ and $CLS_{t+\Delta t}$ of the partially visible future, while the prior ($p_\psi$) is learned from $CLS_t$ and \textit{TE}. After sampling stochastic embedding (\textit{SE}) ($\hat{z}_{t+\Delta t}$) from the posterior $q_\phi$, a cross-attention-based decoder queries $\Tilde{h}_{t+\Delta t}$ against $[\hat{z}_{t+ \Delta t}, h_t]$ to  reconstruct the future visit ($\hat{\vx}_{t+\Delta t}$). Once pretrained, the components within the dashed blue line are available during inference for the downstream task. $\sim\!\!\triangleright$ indicates sampling operation.}
\label{fig_place}
\end{figure*}

\section{Methodology}



STAMP learns enhanced temporal features by improving future visit reconstruction of SiamMAE with conditioning on the time difference between two visit volumes and introducing stochasticity. The time difference conditioning also allows us to time-prompt the model during inference, which can be used for prognostic tasks with varying time windows solely using a single scan. In addition, to enable the network to model multiple possible future outcomes, we introduce time-conditioned stochasticity with a learned prior based on the past visit volume along with time difference and a posterior incorporating the future volume. 


STAMP pretraining pipeline is depicted in Figure~\ref{fig_place}. The following sections describe and elaborate each step of the pipeline in order: \textbf{(1)} input and image masking, \textbf{(2)} temporal encoding, \textbf{(3)} stochasticity, \textbf{(4)} decoding (reconstruction), including \textbf{(5)} loss derivation.

\subsubsection{3D masking strategy} The longitudinal datasets consist of routinely monitored patients with 3D volumes $\vx \in\mathbb{R}^{D\times H \times W}$ and $\vx_t$ denotes the scan acquired at visit day $t$ from $\mathbb{T} = \lbrace t \in \mathbb {N}_0; t \leq b \rbrace$ with $b$ being the last visit, both are expressed in terms of days from the baseline visit. Then, from a patient, two visits are randomly sampled, yielding volumes $\vx_t$ and $\vx_{t+\Delta t}$ with a time interval of $\Delta t \in \{n \in \mathbb{Z} \mid \Delta t_{min} \leq n \leq \Delta t_{max}\}$ between their acquisitions. 
Following the standard vision transformer pipeline, each volume is patchified using a 3D convolutional kernel~\cite{dosovitskiy2021an,chen2023masked} (projection layer) into non-overlapping patches. In line with SiamMAE~\cite{gupta2023siamese} for the SSL task of future reconstruction, 75\%~\cite{kunanbayev2024training, zhou2023retfound} of $\vx_{t+\Delta t}$ is masked, with the remaining visible patches denoted by $\Tilde{\vx}_{t+\Delta t}$. Fixed 3D sin-cos positional encodings are added to the patch embeddings of $\vx_t$ and $\Tilde{\vx}_{t+\Delta t}$. Finally, a learnable CLS token is appended to each input.

\subsubsection{Temporal encoding (TE)} We propose to incorporate time difference information using temporal encodings (\textit{TE}) that are generated similar to diffusion step embedding with in-context setup~\cite{peebles2023scalable}. \textit{TE} is generated by a learnable 2-layer MLP following~\cite{peebles2023scalable} with a SiLU non-linearity function~\cite{elfwing2018sigmoid}, which takes a 1D sin-cos wave from discretized $\Delta t$ as input and embeds it.
\begin{equation}
\begin{split}
H = \tfrac{dim(embedding)}{2},\quad \omega_i = \exp\!\Bigl(-\ln(\mathrm{200})\,\tfrac{i}{H}\Bigr)\;\\
\mathrm{TE}(\Delta t) = \mathrm{MLP}(\bigl[\cos(\Delta t\,\omega_i),\;\sin(\Delta t\,\omega_i)\bigr]_{i=0}^{H-1}).
\end{split}
\end{equation}
Then, \textit{TE} is used similar to the positional encodings as an additive bias to the CLS token before the encoder, similar to~\cite{cong2022satmae}. To be able to prompt the decoder at the inference time, a second \textit{TE} is added before the decoder. In both cases, \textit{TE} is added to the previous visit's $CLS_t$ token (Fig.~\ref{fig_place}). The token is referred as CLS token as it is the nomenclature, even though there is no supervised task in the STAMP pretraining.
Finally, a ViT-based encoder $f_{\theta}$ embeds patch and CLS tokens as $h_t$ and $\Tilde{h}_{t+\Delta t}$ from $\vx_t$ and partially masked $\Tilde{\vx}_{t+\Delta t}$ in a Siamese setup.

\subsubsection{Stochasticity}After the encoder, $CLS_t$ and $CLS_{t+\Delta t}$ are used as stochasticity tokens similar to~\cite{petrovich2021action}. In the Siamese setup, during training, the network has 3 inputs: the fully visible past $\vx_t$, the masked future $\Tilde{\vx}_{t+\Delta t}$, and the time difference $\Delta t$. At inference, the model will have access only to the past volume  $\vx_t$ and the time difference (Fig.~\ref{fig_place} dashed area) for the future progression prediction. As the target for our prior, the posterior is constructed with access to both CLS tokens extracted from the past and the partially visible future volumes ($CLS_t$ and $CLS_{t+\Delta t}$). This diverges from RSP, where a fully visible future $\vx_{t+\Delta t}$ is used, because it has been shown that a fully visible future could lead to posterior collapse to the mode of the future~\cite{schmidt2024learning}. 

STAMP solves this by masking the future, and constructing a time conditioned prior. We aim to learn a prior distribution $p_\psi (z_{t+\Delta t}|h_t, \Delta t)$ that approximates the posterior without access to the future scan, unlike VAEs that use a fixed Gaussian prior or RSP prior $p_\psi (z_{t+\Delta t}|h_t)$ . We condition the prior~\cite{denton2018stochastic,jang2024visual} on  $CLS_t$ and $\Delta t$. Two separate 2-layer MLPs with SiLU activation, $p_{\psi}$ and $q_{\phi}$ (Fig~\ref{fig_place}), generate the prior and posterior distribution logits respectively. Following~\cite{hafner2023mastering}, the outputs of the MLPs are used as logits to one-hot categorical distributions to interpret them as probability distributions to capture latent modes of disease progression. 

The proposed \textbf{Prior} retains the temporal information from the \textit{TE} (added to $CLS_t$) which allows us to interpret the stochasticity as a conditional variational inference. On the other hand, the \textbf{Posterior} $q_\phi (\hat{z}_{t+\Delta t}|h_t, \Tilde{h}_{t + \Delta t})$ is aware of the partial future due to masking ($CLS_{t+\Delta t}$), contrary to RSP. The stochastic embedding (\textit{SE}) $\hat{z}_{t+ \Delta t}$ is obtained by sampling from the posterior, which is concatenated as a token with $h_t$ to be used in the decoder. Straight-through estimator~\cite{bengio2013estimating} is used for sampling during the forward pass to ensure uninterrupted gradient flow. The operation is as follows ($\operatorname{SG}$ indicates stop-gradient operation and $ p_\theta(x)$ indicates probabilities of the logits from the distribution definition):\footnote{Reparametrization trick for Categorical Distribution requires Gumbel-Softmax relaxation}
\begin{equation}
z \sim \mathrm{OneHot}\bigl(p_\theta(x)\bigr), 
\quad
\hat z = z + p_\theta(x) - \operatorname{SG}\bigl(p_\theta(x)\bigr).
\end{equation}


\subsubsection{Decoding} The transformer decoder $p_\varphi$ is based on cross-self-attention~\cite{gupta2023siamese} and aims to construct the masked patches of the future scan. The tokens $[\hat{z}_{t+ \Delta t}, h_t]$ from the past visit are used as $Key$ and $Value$, while the masked future tokens $\Tilde{h}_{t+\Delta t}$ concatenated with learnable $[MASK]$ tokens, are used as $Query$ (Fig.~\ref{fig_place} \textit{Decoder}), unlike a self-attention only decoder where the future tokens  $\Tilde{h}_{t+\Delta t}$ only attend to themselves. Before the decoder, the second \textit{TE} is added to $CLS_t$, while the decoder positional encodings are added to both branches. The decoder output is used in the reconstruction loss for the masked patches.

\subsubsection{Loss function}

STAMP pretraining loss is based on VAE loss. VAEs use evidence lower bound (ELBO) loss to reconstruct realistic images and bring the posterior learned from the latent representation distribution to the prior from a fixed distribution (Gaussian) with the following formula (input image is indicated as $x_t$ for consistency with STAMP):
\begin{equation}
\begin{split}
\mathcal{L}
 =
\mathbb{E}_{z_t\sim q_{\phi}}\bigl[\log p_{\varphi}(x_t\mid z_t)\bigr]
-
\beta\,D_{KL}\!\bigl(q_{\phi}\,\big\|\,\mathcal{N}(0,I)\bigr)
\end{split}
\end{equation}
 Later, VAE was extended to condition the latent space on a class label to increase the likelihood of generating images belonging to the given class (CVAE)~\cite{kingma2014semi}. While the posterior $q_{\phi}$ is conditioned on both the input image $\vx_t$ and the condition $t$, the prior is still a fixed distribution. The condition is added to both the encoder and the decoder. The ELBO loss is: 
\begin{equation}
\begin{split}
\mathcal{L}
 &=
\mathbb{E}_{z_t\sim q_{\phi}}\bigl[\log p_{\varphi}(x_t\mid z_t,t)\bigr] \\
 &-
\beta\,D_{\mathrm{KL}}\!\bigl(q_{\phi}(z_t\mid x_t,t)\,\big\|\,\mathcal{N}(0,I)\bigr)
\end{split}
\end{equation}
Following from CVAE, STAMP adapts its loss to the Siamese setup with masking, where $q_\phi$ is conditioned on both $\vx_t$ and $\Tilde{\vx}_{t+\Delta t}$, while the prior $p_\psi$ is only informed from $\vx_t$ and $\Delta t$. 
Since the sampled variable $z$ is modeled as a latent variable of the disease development, we assume that there is a dependency between $z$ and $\Delta t$, hence the ELBO becomes\footnote{In the network implementation, $\vx$s in the formula are the embeddings $\vh$}:

\begin{equation}
\begin{split}
    \mathcal{L} &= \; \mathbb{E}_{\hat{z}_{t+\Delta t}\sim q_\phi} \Big[
    \underbrace{\log p_\varphi\Big(\mathbf{\hat{\vx}_{t+\Delta t}}
    \mid \mathbf{\hat{z}_{t+ \Delta t}}, \mathbf{x_t}, \mathbf{\Tilde{\vx}_{t+\Delta t}}\Big)}_{\text{reconstruction loss}}
    \Big] \\
    & - \beta\, D_{KL}\Big[\underbrace{ q_\phi\Big(\mathbf{\hat{\vz}_{t+\Delta t}}\mid \mathbf{\vx_t}, \mathbf{\Tilde{\vx}_{t+\Delta t}}\Big)
    \,\Big\|\, p_\psi\Big(\mathbf{\vz_{t+\Delta t}}\mid \mathbf{\vx_t},\mathbf{\Delta t}}_{KL-divergence}\Big)\Big]
\end{split}
\end{equation}

The first term is the masked future reconstruction with MSE loss:
\begin{equation}
\mathcal{L_{MSE}}=\frac{1}{|\mathcal{M}|}\sum_{m \in \mathcal{M}} |\hat{\vx}^{(m)}_{t+\Delta t} - \vx^{(m)}_{t+\Delta t}|^2
\end{equation}
where $m$ is an index from the list of masked patches $\mathcal{M}$.

The second term brings the prior distribution conditioned on the past scan and the future time-interval alone, closer to the posterior distribution which  has access to both the past as well as unmasked parts of the future volume enabling it to learn meaningful temporal representations. During implementation, a symmetric KL-divergence loss with \textit{stop-gradient}~\cite{hafner2023mastering} was used for numerical stability. Hafner et al.~\cite{hafner2023mastering} showed that stop gradient and larger bias for the posterior are crucial for preventing posterior collapse. Similar regularization is employed in~\cite{jang2024visual,burchi2025accurate}. The full implementation of $D_{KL}$ is:
\begin{equation}\label{eq:kl}
\begin{split}
 D_{KL} = 0.2 *  KL\Big[ q_\phi
    \,\Big\|\, \operatorname{SG}(p_\psi)\Big] \\ +
0.8 *  KL\Big[ \operatorname{SG}(q_\phi)
    \,\Big\|\, p_\psi\Big]    
\end{split}
\end{equation}

\section{Experimental Setting}

\subsection{Datasets}  The HARBOR\footnote{NCT00891735. https://clinicaltrials.gov/ct2/show/NCT00891735} dataset consists of 1076 patients (eyes) with AMD in different stages, with monthly OCT scans over a 24 month period using Cirrus OCT scanner. We split the data into two separate parts for pretraining and wet-AMD conversion downstream prediction task with 5\% converted within 6 months (Table~\ref{Tbl:stat} HARBOR). We further divided the downstream data into 4 folds at the patient level for cross-validation and reported the average results. To evaluate under image domain-shift and an additional task, we employed an external dataset, part of the retrospective PINNACLE initiative~\cite{sutton2023developing}, consisting of scans from Topcon device with labels for conversion to GA from iAMD, within 12 months (Table~\ref{Tbl:stat} PINNACLE). It contains 550 patients (eyes) with 4260 OCT scans of which 7\% converted within 12 months. OCT spatial dimensions are resized to $448\times448$, and we used the central 32 B-Scans (1.5 mm) as the volumetric input. 

\begin{figure}[tbh]
    \centering
    \includegraphics[width=\linewidth]{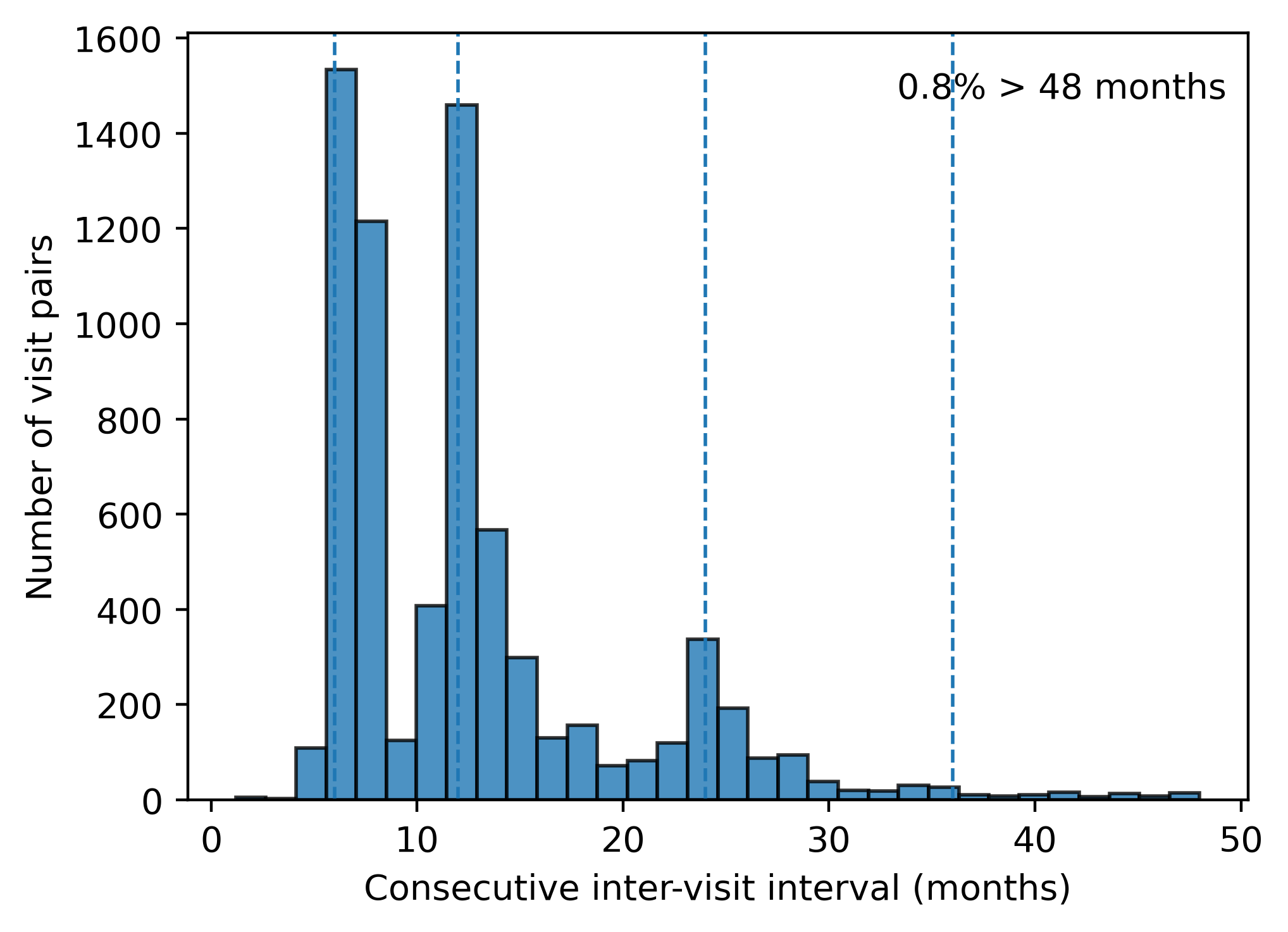}
\caption{Intra-visit intervals of ADNI dataset. Dashed lines indicate 1-, 2-, and 3-years.} 
\label{fig:adni_interv}
\end{figure}

We also used the ADNI~\cite{petersen2010alzheimer} cohort for 3D brain MRI experiments\footnote{Data used in the preparation of this article were obtained from the Alzheimer’s Disease Neuroimaging Initiative (ADNI) database (adni.loni.usc.edu). The ADNI was launched in 2003 as a public-private partnership, led by Principal Investigator Michael W. Weiner, MD. The primary goal of ADNI has been to test whether serial magnetic resonance imaging (MRI), positron emission tomography (PET), other biological markers, and clinical and neuropsychological assessment can be combined to measure the progression of mild cognitive impairment (MCI) and early Alzheimer’s disease (AD)}.  The subset we used has 9265 T1-weighted MRIs with 3344 CN, 4022 MCI, and 1899 AD cases from 1990 patients (Table~\ref{Tbl:stat} ADNI). Following~\cite{puglisi2024enhancing}, the preprocessing steps with the order are Bias Field Correction~\cite{tustison2010n4itk}, skull stripping~\cite{hoopes2022synthstrip}, rigid registration to the 1 mm isotropic MNI template~\cite{avants2008symmetric} yielding $182\times218\times182$ sized images. 
Finally, a center crop of $128\times128\times128$ is applied. The downstream task data consists of the MCI diagnosed volumes evaluated for conversion to AD within 1-3 years as a binary label. Unlike HARBOR, ADNI has highly irregular visit intervals (Table~\ref{Tbl:stat} and Figure~\ref{fig:adni_interv})

\setlength{\tabcolsep}{3pt}
\begin{table}[!t]
  \centering
    \caption{Statistics of the datasets. Disease pathways are defined as a detected conversion during the study.}\label{Tbl:stat}
    \begin{tabular}{llccc}
      \toprule
      Dataset  & Class  & \# of Patients & \# of Visits & Interval (months) \\
      \midrule
      \multirow{3}{*}{HARBOR}      & pretraining  & 579            & 12770      & \multirow{3}{*}{$1.0\pm1.0$} \\
                                   & iAMD$\rightarrow$iAMD  &  431        & 8141      &  \\
                                   & iAMD$\rightarrow$wet-AMD  & 117            & 1967      &  \\
      \midrule
      \multirow{2}{*}{PINNACLE}    & iAMD$\rightarrow$iAMD      &451            & 4260      & \multirow{2}{*}{$3.2\pm3.3$}  \\
                                   & iAMD$\rightarrow$GA  & 99            & 621      & $ $ \\
      \midrule
      \multirow{4}{*}{ADNI}
               & CN$\rightarrow$CN       & 584            & 2665        & \multirow{4}{*}{$12.9\pm8.5$}   \\
               & MCI$\rightarrow$MCI     & 569            & 2570       &  \\
               & MCI (CN)$\rightarrow$AD     & 358            & 2112        & \\
               & AD       & 332            &1043        & \\
      \bottomrule
    \end{tabular}
\end{table}

\setlength{\tabcolsep}{4pt}
\begin{table*}[tbh]
\centering
\caption{Ablation for the proposed temporal (TE) and stochastic (SE) components during pretraining and the downstream task on the HARBOR dataset. Row 7 corresponds to STAMP. In the Pretraining column, (\xmark) indicates a method without the component. In the Downstream column, (\xmark) indicates a component added in the pretraining but not used in the downstream task. PRAUC baselines for 6 and 12 months are 0.05 and 0.11, respectively.}

\label{Tbl:AblationPretraining}
\fontsize{8}{10}\selectfont
\begin{tabular}{l c c c c c c c c c c}
\toprule
  & \multicolumn{2}{c}{\textbf{Pretraining}}
  & \multicolumn{2}{c}{\textbf{Downstream}}
  & \multicolumn{3}{c}{\textbf{6-months}}
  & \multicolumn{3}{c}{\textbf{12-months}} \\
\cmidrule(lr){2-3} \cmidrule(lr){4-5} \cmidrule(lr){6-8} \cmidrule(lr){9-11}
  &  SE &  TE 
  &  SE &  TE 
  & AUROC ↑ & PRAUC ↑ & BACC ↑ 
  & AUROC ↑ & PRAUC ↑ & BACC ↑ \\
\midrule
1         
  &  \xmark      & \checkmark        
  &     -   & \xmark
  & $0.634\pm0.019$ & $0.111\pm0.019$ & $0.587\pm0.012$  
  & $0.602\pm0.064$ & $0.156\pm0.039$ & $0.564\pm0.052$ \\
2               
  &  \xmark      & \checkmark        
  &   -     &          \checkmark
  & $0.675\pm0.043$ & $0.120\pm0.028$ & \underline{$0.612\pm0.036$}  
  & $0.640\pm0.053$ & $\underline{0.188\pm0.041}$ & $0.591\pm0.033$ \\
3             
  &   \checkmark     &  \xmark       
  & \xmark   &  -     
  & $0.637\pm0.048$ & $0.100\pm0.017$ & $0.592\pm0.028$ 
  & $0.601\pm0.038$ & $0.138\pm0.011$ & $0.571\pm0.033$ \\
4                  
  &   \checkmark     &   \xmark      
  &    \checkmark    &     -     
  & $0.627\pm0.033$ & $0.088\pm0.015$ & $0.579\pm0.016$  
  & $0.602\pm0.028$ & $0.134\pm0.012$ & $0.561\pm0.024$ \\
\midrule
5             
  &   \checkmark      &    \checkmark      
  & \xmark & \checkmark     
  & $\underline{0.690\pm0.036}$ & $\underline{0.127\pm0.027}$ & $0.612\pm0.043$ 
  & $\underline{0.652\pm0.061}$ & $0.180\pm0.063$ & $\underline{0.601\pm0.052}$ \\
6            
  &    \checkmark    &    \checkmark     
  &  \checkmark      & \xmark
  & $0.639\pm0.031$ & $0.108\pm0.022$ & $0.592\pm0.064$ 
  & $0.594\pm0.034$ & $0.128\pm0.027$ & $0.545\pm0.050$ \\
7                     
  &    \checkmark    &  \checkmark       
  &  \checkmark      &          \checkmark
  & $\mathbf{0.700\pm0.047}$ & $\mathbf{0.140\pm0.042}$ & $\mathbf{0.615\pm0.049}$ 
  & $\mathbf{0.671\pm0.057}$ & $\mathbf{0.200\pm0.066}$ & $\mathbf{0.607\pm0.042}$ \\
\bottomrule
\end{tabular}
\end{table*}

\setlength{\tabcolsep}{4pt}
\begin{table*}[tbh]
\centering
\caption{Ablation for the proposed temporal (TE) and stochastic (SE) components during pretraining and the downstream task on the ADNI dataset. Row 7 corresponds to STAMP. In the Pretraining column, (\xmark) indicates a method without the component. In the Downstream column, (\xmark) indicates a component added in the pretraining but not used in the downstream task. PRAUC baselines for 1 and 3 years are 0.07 and 0.18, respectively.}

\label{Tbl:AblationPretrainingADNI}
\fontsize{8}{10}\selectfont
\begin{tabular}{l c c c c c c c c c c}
\toprule
  & \multicolumn{2}{c}{\textbf{Pretraining}}
  & \multicolumn{2}{c}{\textbf{Downstream}}
  & \multicolumn{3}{c}{\textbf{1-year}}
  & \multicolumn{3}{c}{\textbf{3-years}} \\
\cmidrule(lr){2-3} \cmidrule(lr){4-5} \cmidrule(lr){6-8} \cmidrule(lr){9-11}
  &  SE &  TE 
  &  SE &  TE 
  & AUROC ↑ & PRAUC ↑ & BACC ↑ 
  & AUROC ↑ & PRAUC ↑ & BACC ↑ \\
\midrule
1         
  &  \xmark      & \checkmark        
  &     -   & \xmark
  & $0.751\pm0.002$ & $0.195\pm0.002$ & $0.699\pm0.002$  
  & $0.746\pm0.001$ & $0.372\pm0.001$ & $0.696\pm0.005$ \\
2               
  &  \xmark      & \checkmark        
  &   -     &          \checkmark
  & $0.779\pm0.001$ & $0.214\pm0.002$ & \underline{$0.702\pm0.012$}  
  & $0.762\pm0.006$ & $0.362\pm0.016$ & $0.665\pm0.019$ \\
3             
  &   \checkmark     &  \xmark       
  & \xmark   &  -     
  & $0.770\pm0.006$ & $\mathbf{0.300\pm0.014}$ & $0.692\pm0.005$ 
  & \underline{$0.795\pm0.002$} & \underline{$0.438\pm0.005$} & $\mathbf{0.714\pm0.002}$ \\
4                  
  &   \checkmark     &   \xmark      
  &    \checkmark    &     -     
  & $0.787\pm0.001$ & $0.248\pm0.004$ & $0.698\pm0.006$
  & $\mathbf{0.799\pm0.002}$ & $\mathbf{0.460\pm0.002}$ & \underline{$0.710\pm0.016$} \\
\midrule
5             
  &   \checkmark      &    \checkmark      
  & \xmark & \checkmark     
  & $\underline{0.799\pm0.001}$ & $0.269\pm0.002$ & $0.698\pm0.003$ 
  & $0.771\pm0.007$ & $0.398\pm0.021$ & $0.693\pm0.010$ \\
6            
  &    \checkmark    &    \checkmark     
  &  \checkmark      & \xmark
  & $0.768\pm0.002$ & $0.194\pm0.002$ & $0.676\pm0.007$ 
  & $0.756\pm0.005$ & $0.342\pm0.008$ & $0.689\pm0.026$ \\
7                     
  &    \checkmark    &  \checkmark       
  &  \checkmark      &          \checkmark
  & $\mathbf{0.812\pm0.001}$ & $\underline{0.277\pm0.004}$ & $\mathbf{0.736\pm0.009}$ & $0.773\pm0.001$ &$0.416\pm0.013$ & $0.691\pm0.001$  \\
\bottomrule
\end{tabular}
\end{table*}

\subsection{Baselines}  STAMP pretraining is tested against MAE-based methods: MAE~\cite{he2022masked}, SiamMAE~\cite{gupta2023siamese}, CropMAE~\cite{eymael2025efficient} and RSP~\cite{jang2024visual}. Additionally, we compared against common pretrained weights from ImageNet and, in the case of OCT datasets, MAE-pretrained OCT foundation models RetFound~\cite{zhou2023retfound} and VisionFM~\cite{visionfm}. Both models were pretrained on 2D B-scans of OCT volumes. Thus, we expanded the initial projection layer by inflating the convolutional kernel~\cite{carreira2017quo}. Then, we provided extensive ablation studies. We used AMD progression prediction task on two datasets for evaluation on frozen backbones. For the disease progression, only the future is relevant, thus we kept the $\Delta t$ positive between pairs and only considered future OCT reconstructions. We also provided pretraining baseline comparisons on 3D MRIs from ADNI cohort for Alzheimer's Disease progression.

\subsection{Pretraining setting}  We used ViT-Base as the encoder (12 blocks with embedding dimension of 768). The decoder consists of 6 blocks of cross-attention self-attention pairs with an embedding dimension of 384. Following~\cite{hafner2023mastering}, CLS tokens are split into 32 dimensional bins to be used in the straight-through estimator for stochasticity. All methods were pretrained for 800 epochs, with a batch size of 96, a learning rate of $1e-4$, and a weight decay of $1e-2$ in AdamW~\cite{loshchilov2018decoupled}. OCT pairs are randomly sampled with a $\Delta t$ of $3-18$ months and a monthly discretization for \textit{TE}. OCT inputs are size of $32\times448\times448$ and patchified into patches of $4\times32\times32$. In ADNI experiments, the MRIs are of size $128\times128\times128$ and patchified into patches of $16\times16\times16$. $\Delta t$ is calculated with a discretization of 6 months. We used random horizontal flip, color jittering, and cropping. We found that horizontal flip is crucial for preventing the network from ignoring the future and reconstructing solely based on the past in line with the results reported by Eyma{\"e}l et al.~\cite{eymael2025efficient}.  All experiments were run on a single NVIDIA A100 GPU.

\subsection{Downstream task evaluation} After pretraining, the models were evaluated with frozen backbones and employed pretrained TE \& SE, if available, for highlighting the contribution of each component. The task is a binary conversion prediction given a time window (6-12 months for HARBOR, 12 months for PINNACLE, 1-3 years for ADNI) with the Cross-Entropy loss. We used Attention Pooling~\cite{chen2024context, elscalable} as an alternative to linear evaluation. After obtaining the input features from the encoder, Attention Pooling consists of a single cross-attention layer added with an additional CLS token to attend the patch embeddings relevant for classification without any non-linearity. 
All experiments were run for 200 epochs with Adam optimizer. Learning rates are set using grid-search over the validation splits. We reported results with area under the receiver operating characteristic (AUROC), area under the precision-recall curve (PRAUC), and balanced accuracy (BACC), which corresponds to the average recall of classes. While the random prediction baseline for AUROC and BACC is $0.5$, in PRAUC it is proportional to the positive class ratio. The PRAUC baselines are: 0.05 for HARBOR 6 months, 0.11 for HARBOR 12 months, 0.07 for PINNACLE 12 months, 0.07 for ADNI 1 year and 0.18 for ADNI 3 years tasks. 

Additionally, we provided a wet-AMD risk estimation model for the intermediate stage patients in HARBOR using a Cox model with time varying covariates~\cite{zhang2018time,andersen1982cox}. We reported Concordance index (c-index) results based on the predicted risk of each visit. Similar to the conversion prediction, a trainable attention pooling head aggregates OCT token embeddings to a survival head that outputs a scalar log-risk score. Following neural extensions of Cox regression that allow the relative risk to depend explicitly on time~\cite{kvamme2019time}, the model outputs a visit-specific log-risk score:
\begin{equation}\label{eq:risk}
r(t) = g(h_t, t)
\end{equation}

To account for multiple visits per patient and right censoring, we use the time-varying Cox/counting-process formulation~\cite{zhang2018time,andersen1982cox}. Each visit defines an interval $(s_i,e_i]$, where the OCT representation from time $s_i$ is active until the next visit, conversion, or censoring. The model is trained with the negative partial log-likelihood:
\begin{equation}
\begin{split}
&\mathcal{L}
= -\sum_{i:\delta_i=1}\left[r_i -\log \sum_{j\in R(e_i)} \exp(r_j)\right], \\
&R(e_i)=\{j:s_j<e_i\le e_j\}
\end{split}
\end{equation}
Here, $\delta_i$ indicates whether conversion occurred at the end of interval $i$. Treatment-response modeling requires treatment metadata, dosing schedules, and outcome definitions that are not consistently available across the datasets used in this study. We therefore leave treatment-response prediction as future work.

\section{Results}

\setlength{\tabcolsep}{3pt}
\begin{table*}[tbh] 
\centering
\caption{Comparison of time encoding as an additive bias, i.e., the original STAMP design denoted as STAMP$_{embedding}$, and as a separate input token, denoted as STAMP$_{token}$. Results are reported over 4 folds on HARBOR and over repeated experiments on ADNI.}
\label{Tbl:token}
\fontsize{8}{10}\selectfont
\begin{tabular}{c l c c c c c c}
\toprule
\multirow{4}{*}[-10pt]{\rotatebox[origin=c]{90}{\textbf{HARBOR}}}
& \multirow{2}{*}{\textbf{Model}}
  & \multicolumn{3}{c}{\textbf{6-months}}
  & \multicolumn{3}{c}{\textbf{12-months}} \\
\cmidrule(lr){3-5} \cmidrule(lr){6-8}
& & AUROC $\uparrow$ & PRAUC $\uparrow$ & BACC $\uparrow$
  & AUROC $\uparrow$ & PRAUC $\uparrow$ & BACC $\uparrow$ \\
\cmidrule(lr){2-8}
& STAMP$_{embedding}$
  & $0.700\pm0.047$ & $0.140\pm0.042$ & $0.615\pm0.049$
  & $0.671\pm0.057$ & $0.200\pm0.066$ & $0.607\pm0.042$ \\
& STAMP$_{token}$
  & $0.694\pm0.035$ & $0.139\pm0.033$ & $0.620\pm0.030$
  & $0.674\pm0.045$ & $0.198\pm0.051$ & $0.607\pm0.037$ \\
\midrule\midrule
\multirow{3}{*}[-6pt]{\rotatebox[origin=c]{90}{\textbf{ADNI}}}
&
  & \multicolumn{3}{c}{\textbf{1-year}}
  & \multicolumn{3}{c}{\textbf{3-years}} \\
\cmidrule(lr){3-5} \cmidrule(lr){6-8}
& STAMP$_{embedding}$
  & $0.812\pm0.001$ & $0.277\pm0.004$ & $0.736\pm0.009$
  & $0.773\pm0.001$ & $0.416\pm0.013$ & $0.691\pm0.001$ \\
& STAMP$_{token}$
  & $0.793\pm0.002$ & $0.211\pm0.001$ & $0.762\pm0.011$
  & $0.796\pm0.002$ & $0.371\pm0.001$ & $0.732\pm0.008$ \\
\bottomrule
\end{tabular}
\end{table*}

\setlength{\tabcolsep}{3pt}
\begin{table}[h]
\centering
\caption{Ablation studies for Masking ratio (75\% and 25\%) and symmetric KL term weighting (0.2--0.8 and 0.8--0.2 for posterior-prior respectively). The models are pretrained with the ADNI dataset.}\label{Tbl:maskabl}
{\fontsize{8}{10}\selectfont
\begin{tabular}{lccc}
\toprule
\textbf{Model} & \textbf{AUROC $\uparrow$} & \textbf{PRAUC $\uparrow$} & \textbf{BACC $\uparrow$} \\ \midrule
0.8-0.2        & $0.788\pm0.003$ & $0.258\pm0.002$ &$0.723\pm0.013$\\
25\%       & $0.778\pm0.002$ & $0.251\pm0.008$ &$0.718\pm0.014$\\
\hline  
STAMP        & $\mathbf{0.812\pm0.001}$ & $\mathbf{0.277\pm0.004}$ &$\mathbf{0.736\pm0.009}$\\
\bottomrule
\end{tabular}}
\end{table}

\setlength{\tabcolsep}{3pt}
\begin{table*}[tbh]
\centering
\caption{Linear evaluation of wet-AMD conversion prediction on pretrained models within two time‐windows: 6 and 12 months on the HARBOR Dataset. ImageNet, VisionFM and RetFound are fully finetuned. The results are reported over 4-folds}
\label{Tbl:Results}
\fontsize{8}{10}\selectfont
\begin{tabular}{l c c c c c c}
\toprule
\multirow{2}{*}{\textbf{Model}}  
  & \multicolumn{3}{c}{\textbf{6-months}} 
  & \multicolumn{3}{c}{\textbf{12-months}} \\
\cmidrule(lr){2-4} \cmidrule(lr){5-7}
  & AUROC $\uparrow$ & PRAUC $\uparrow$ & BACC $\uparrow$  
  & AUROC $\uparrow$ & PRAUC $\uparrow$ & BACC $\uparrow$ \\
\midrule
VisionFM\cite{visionfm}    
  & $0.631\pm0.036$ & $0.090\pm0.010$ & $0.583\pm0.075$  
  & $0.593\pm0.053$ & $0.126\pm0.017$ & $0.553\pm0.058$ \\
ImageNet                    
  & $0.665\pm0.029$ & $0.096\pm0.026$ & $\underline{0.614\pm0.033}$  
  & $\underline{0.640\pm0.022}$ & $0.151\pm0.052$ & $0.578\pm0.013$ \\
RetFound\cite{zhou2023retfound} 
  & $0.636\pm0.043$ & $0.091\pm0.027$ & $0.560\pm0.024$         
  & $0.551\pm0.024$ & $0.121\pm0.042$ & $0.531\pm0.033$ \\
\midrule
TC\cite{emre2024learning}      
  & $0.643\pm0.033$ & $0.101\pm0.011$ & $0.611\pm0.050$ 
  & $0.599\pm0.067$ & $0.143\pm0.042$ & $0.564\pm0.057$ \\
MAE\cite{he2022masked}      
  & $0.620\pm0.027$ & $0.087\pm0.014$ & $0.562\pm0.025$ 
  & $0.602\pm0.056$ & $0.137\pm0.033$ & $0.531\pm0.029$ \\
SiamMAE\cite{gupta2023siamese}   
  & $0.666\pm0.022$ & $\underline{0.108\pm0.016}$ & $0.606\pm0.024$ 
  & $0.638\pm0.038$ & $\underline{0.161\pm0.036}$ & $\underline{0.590\pm0.038}$ \\
CropMAE\cite{eymael2025efficient}    
  & $\underline{0.675\pm0.020}$ & $0.105\pm0.017$ & $0.612\pm0.023$ 
  & $0.523\pm0.018$ & $0.097\pm0.011$ & $0.503\pm0.008$ \\
RSP\cite{jang2024visual}      
  & $0.580\pm0.035$ & $0.084\pm0.025$ & $0.532\pm0.026$ 
  & $0.540\pm0.044$ & $0.113\pm0.032$ & $0.521\pm0.033$ \\
\midrule
STAMP                        
  & $\mathbf{0.700\pm0.047}$ & $\mathbf{0.140\pm0.042}$ & $\mathbf{0.615\pm0.049}$ 
  & $\mathbf{0.671\pm0.057}$ & $\mathbf{0.200\pm0.066}$ & $\mathbf{0.607\pm0.042}$ \\
\bottomrule
\end{tabular}
\end{table*}

\subsection{Effect of Temporal Encoding and Stochasticity}

We conducted ablation studies to show the importance of pretraining with temporal encodings and stochasticity. Until now, most SSL-based models discarded the extra components that are learned during the pretraining, focusing on the general representation quality. We hypothesized that these components learn meaningful latent factors that could be lost due to fine-tuning or just discarding. We, therefore performed the ablation in two folds: adding TE \& SE during pretraining, and employing them during the downstream evaluation (Fig.~\ref{fig_place} blue dashed area). We provided the proposed ablation setup in both HARBOR (Table~\ref{Tbl:AblationPretraining}) and ADNI (Table~\ref{Tbl:AblationPretrainingADNI}) datasets.

First, we added unconditional stochasticity on SiamMAE (Table~\ref{Tbl:AblationPretraining} Row 3). If RSP~\cite{jang2024visual} is simplified without its third encoder pass (alongside the second reconstruction) and with masking the future input, it would yield the same model. Compared to RSP (Table~\ref{Tbl:Results}), reducing the pipeline complexity and the number of loss terms, clearly improved the pretraining stability, as evidenced by improvements across all metrics and tasks. We tested the learned stochasticity by sampling a stochastic token (\textit{SE}) from its unconditional prior $z_t\sim p_\psi (z_{t}|h_t)$, and concatenated it ($[z_{t}, h_t]$) (Table~\ref{Tbl:AblationPretraining} Row 4) during the downstream evaluation. Due to its prior being unaware of $\Delta t$, it led to a decrease in the performance, highlighting the need for a \textit{TE}. Similarly, low performance of RSP confirms our hypothesis that long time-intervals typical of medical imaging require time conditioning. This is supported by SiamMAE outperforming RSP. Thus, it can be concluded that stochasticity without conditioning increases ambiguity in the feature space, degrading the model's performance.




  
  

  
  

Next, we introduced temporal encodings on top of SiamMAE during pretraining (Table~\ref{Tbl:AblationPretraining} Row 1-2). If the trained weights were transferred to ViT3D for the downstream task without employing \textit{TE} during the downstream evaluation (Table~\ref{Tbl:AblationPretraining} Row 1), it actually underperformed compared to SiamMAE (Table~\ref{Tbl:Results}). We speculate that some of the temporal latent factors are learned by \textit{TE}. To demonstrate this, during the inference we prompted the network with $\Delta t$ of 6 and 12 months for each task using the pretrained \textit{TE}. The performance gain (Table~\ref{Tbl:AblationPretraining} Row 2) compared to not using \textit{TE} (Table~\ref{Tbl:AblationPretraining} Row 1) during the inference and SiamMAE supports our claim. Finally, the combination of the aforementioned two components yields the proposed STAMP (Table~\ref{Tbl:AblationPretraining} Row 7), which achieved the best results among all HARBOR ablation configurations across all metrics. Even if SE is not employed during the downstream evaluation (Table~\ref{Tbl:AblationPretraining} Row 5), it achieved the second-best AUROC for both time windows, underscoring the quality of the pretraining phase.

On ADNI (Table~\ref{Tbl:AblationPretrainingADNI}), the 1-year task follows the same trend as HARBOR. Employing a pretrained component during the downstream evaluation improves AUROC and BACC in every case (Rows 1$\rightarrow$2, 3$\rightarrow$4, 5$\rightarrow$7, 6$\rightarrow$7). STAMP (Row 7) achieves the best AUROC and BACC, while Row 3 achieves a higher PRAUC. The 3-year task behaves differently. The models pretrained with SE but without TE (Rows 3--4) achieve the best results on all three metrics, and STAMP ranks third in AUROC and PRAUC. Considering the pretraining setups (Rows 2, 4, 7), STAMP achieves the second best results in the 3-year task. This is in line with Table~\ref{Tbl:ADNI_Results}, where SiamMAE, which also has no TE, achieves the best 3-year AUROC and BACC.

To further investigate how temporal information interacts with spatial image features, we compared the proposed additive temporal encoding strategy, denoted as STAMP$_{embedding}$, with an alternative design in which the temporal embedding is introduced as a separate input token, denoted as STAMP$_{token}$. This experiment directly evaluates whether an explicit temporal token is necessary for temporal-spatial interaction, or whether adding TE to the CLS token is sufficient. As shown in Table~\ref{Tbl:token}, both designs achieved comparable performance across HARBOR and ADNI, indicating that the additive TE is not merely a static bias but effectively interacts with image tokens through transformer attention. On HARBOR, STAMP$_{embedding}$ and STAMP$_{token}$ performed almost identically. On ADNI, STAMP$_{token}$ improved BACC for both 1-year and 3-year prediction, while STAMP$_{embedding}$ achieved higher PRAUC in both settings, which is particularly relevant under class imbalance. Overall, these results suggest that explicit tokenization of TE is not required to obtain temporal-spatial interaction. Therefore, the additive TE design is retained as the default STAMP implementation because it is simpler, introduces no additional sequence token, and achieves competitive or better performance on most metrics.

We further evaluated STAMP's sensitivity to the masking ratio and KL balancing on ADNI (Table~\ref{Tbl:maskabl}). Lower masking decreased performance, suggesting that high masking is important for forcing the model to rely on temporal information from the past scan rather than visible future patches. Similarly, swapping the KL weights (Eq.~\ref{eq:kl}) from 0.2--0.8 to 0.8--0.2 reduced the performance, supporting the use of a larger weight on the prior term for learning an inference-aligned time-conditioned prior.

\setlength{\tabcolsep}{3pt}
\begin{table}[h]
\centering
\caption{Linear evaluation of GA conversion prediction within a 12 months time window on the PINNACLE Dataset. The models are pretrained with the HARBOR dataset. The results are reported over 5-folds}\label{Tbl:DA}
{\fontsize{8}{10}\selectfont

\begin{tabular}{lccc}
\toprule
\textbf{Model} & \textbf{AUROC $\uparrow$} & \textbf{PRAUC $\uparrow$} & \textbf{BACC $\uparrow$} \\ \midrule
TC\cite{emre2024learning}      &0.770 $\pm0.058$ & $0.150\pm0.054$ & $0.702\pm0.046$ \\
MAE\cite{he2022masked}         & $0.830\pm0.025$ & $0.208\pm0.042$ & $0.752\pm0.046$ \\
SiamMAE\cite{gupta2023siamese}   & $\underline{0.833\pm0.036}$  & $\underline{0.229\pm0.053}$  &$\underline{0.755\pm0.040}$\\
CropMAE\cite{eymael2025efficient}        & $0.818\pm0.026$ & $0.174\pm0.041$ & $0.744\pm0.023$ \\
RSP\cite{jang2024visual}         & $0.769\pm0.031$ & $0.155\pm0.036$ & $0.690\pm0.039$ \\ \hline  
STAMP        & $\mathbf{0.848\pm0.020}$ & $\mathbf{0.240\pm0.061}$ &$\mathbf{0.769\pm0.047}$\\
\bottomrule
\end{tabular}}
\end{table}

\subsection{AMD Progression}

STAMP pretraining with TE \& SE employed in the downstream tasks outperformed all baselines across all metrics (Table~\ref{Tbl:Results}), indicating its ability to learn multi-outcome temporal features. Additionally, we included available foundation models, namely VisionFM~\cite{visionfm} and RetFound~\cite{zhou2023retfound}. Both models failed to train when linearly evaluated, thus we fine-tuned them end-to-end. Although they performed better than linearly evaluated MAE, their lack of temporal information and possible domain shift due to scanner type or patient demographics, caused them to underperform significantly compared to STAMP (Table~\ref{Tbl:Results}).

In Table~\ref{Tbl:DA}, we used the pretrained weights from the HARBOR dataset for predicting GA conversion task on the PINNACLE dataset. STAMP achieved the best performance compared to the baselines, highlighting its feature robustness to image domain shift. Also, when varying the amount of labeled HARBOR training data while keeping the STAMP encoder frozen, STAMP showed stable performance across label fractions, indicating that the learned temporal representations remain useful under limited downstream supervision (Table~\ref{Tbl:prop}).

\setlength{\tabcolsep}{3pt}
\begin{table}[h]
\centering
\caption{Label-efficiency analysis for 6-month wet-AMD conversion prediction on HARBOR using different portions of the labeled downstream training data. The pretrained STAMP encoder is frozen and only the downstream classifier is trained. Results are reported over 4 folds.}\label{Tbl:prop}
{\fontsize{8}{10}\selectfont
\begin{tabular}{lccc}
\toprule
\textbf{Model} & \textbf{AUROC $\uparrow$} & \textbf{PRAUC $\uparrow$} & \textbf{BACC $\uparrow$} \\ \midrule
10\%       & $0.679\pm0.016$ & $0.111\pm0.025$ &$0.624\pm0.015$\\
25\%       & $0.691\pm0.044$ & $0.140\pm0.025$ &$0.617\pm0.015$\\
50\%       & $0.719\pm0.034$ & $0.140\pm0.024$ &$0.642\pm0.037$\\
\hline  
100\%        & $0.700\pm0.047$ & $0.140\pm0.042$ &$0.615\pm0.049$\\
\bottomrule
\end{tabular}}
\end{table}

\begin{figure*}[tbh]
\centering
\includegraphics[width=.85\textwidth]{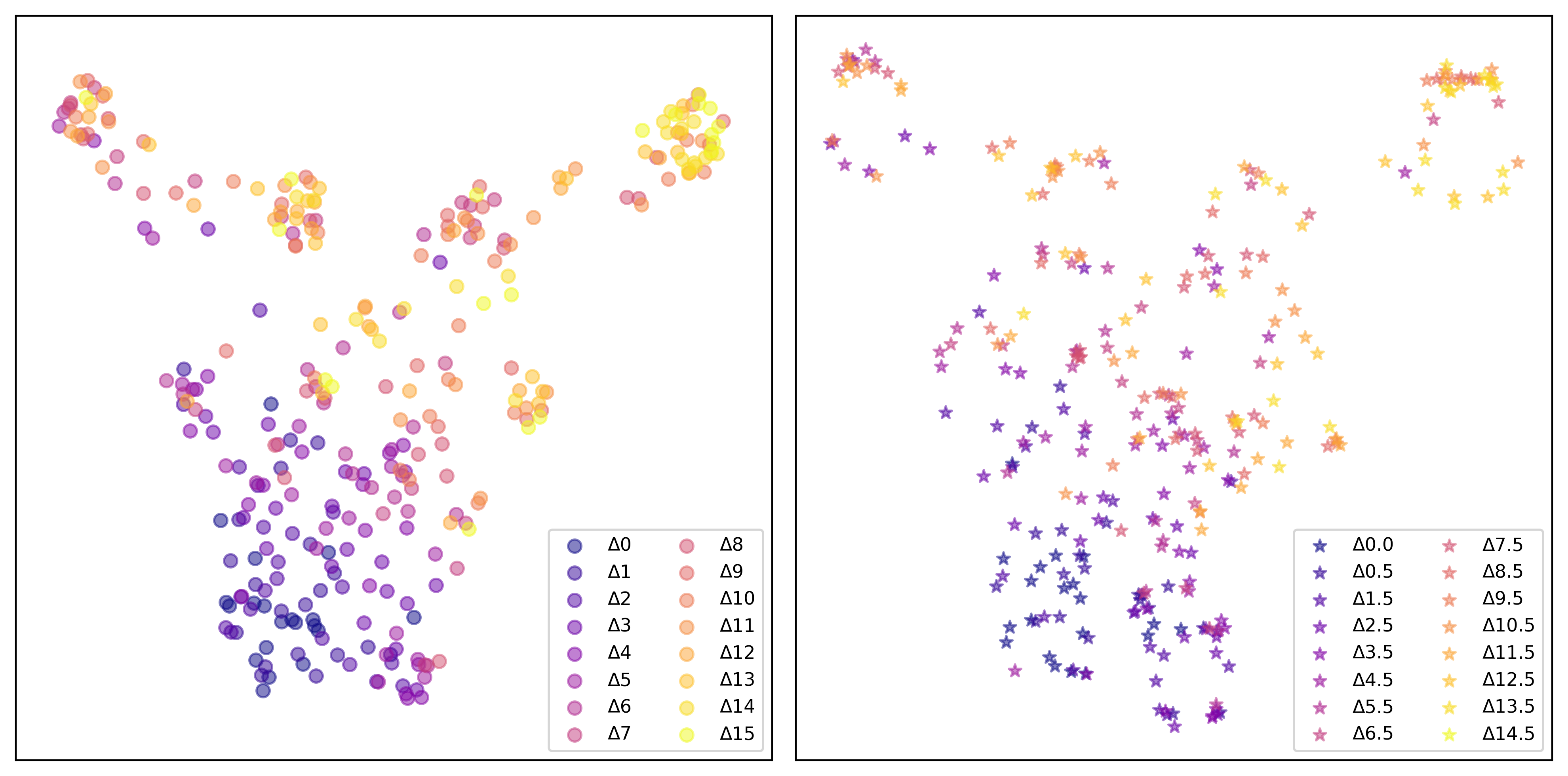}
\caption{\textbf{Left}:  PCA projection of 20 sampled tokens from the learned prior $p_\psi$ per discrete $\Delta t$. \textbf{Right}: PCA projection of 20 sampled tokens from the learned prior $p_\psi$ per fractional $\Delta t$. The temporal alignment indicates diverse but time sensitive stochastic sampling in both cases.} \label{fig1}
\end{figure*}

\begin{figure}[h]
  \centering

  \includegraphics[width=\linewidth]{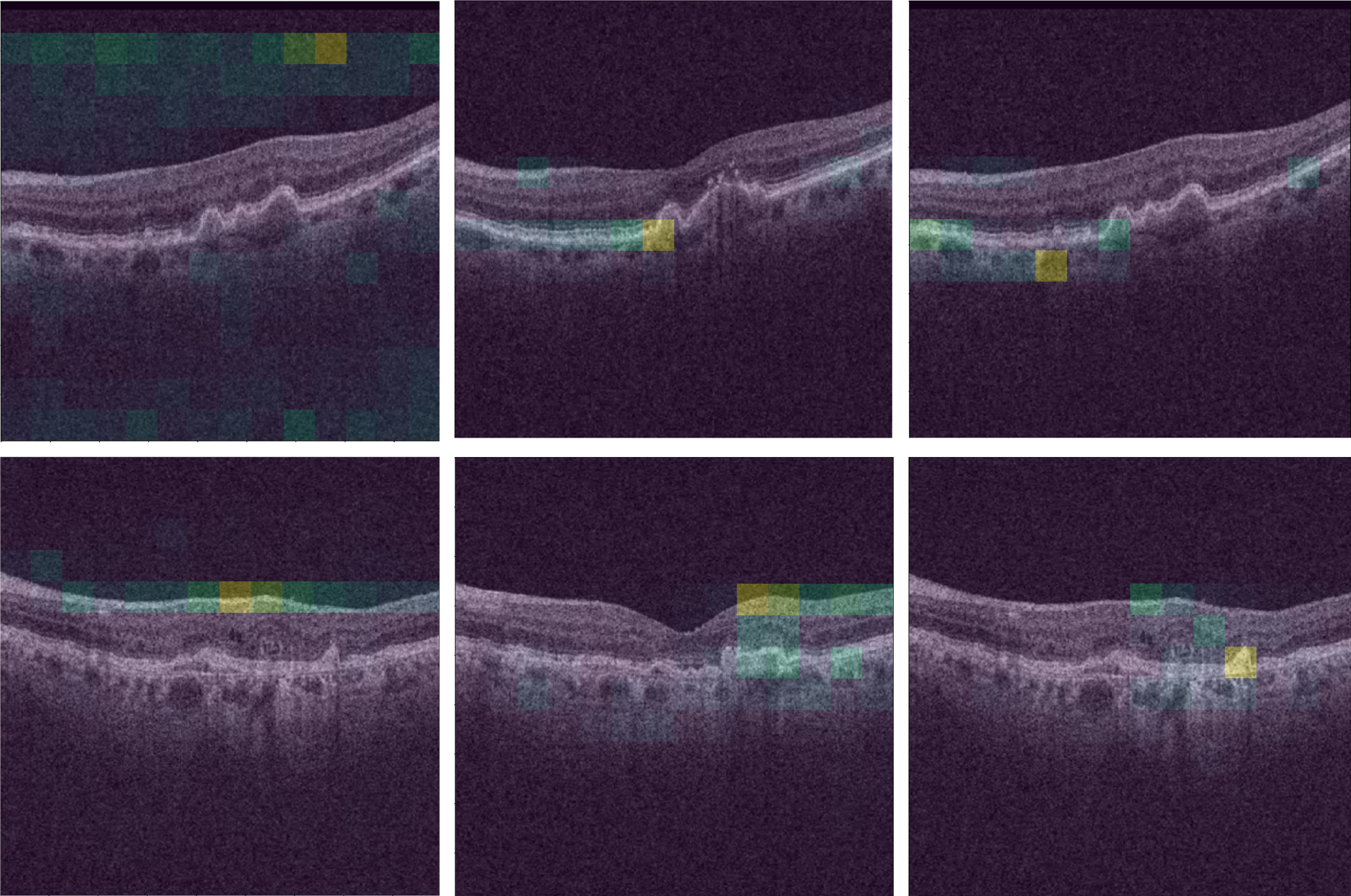}
  
  \caption{Attention map visualization of STAMP w.r.t pretrained CLS. Each row is from a different volume. \textbf{Left:} Attention maps without using \textit{TE}. \textbf{Middle:} Attention maps with 3 months prompted \textit{TE}. \textbf{Right:} Attention maps with 15 months prompted \textit{TE}. }
  \label{fig:attmap}
\end{figure}

\begin{figure}[tbh] 
    \centering
    \includegraphics[width=\linewidth]{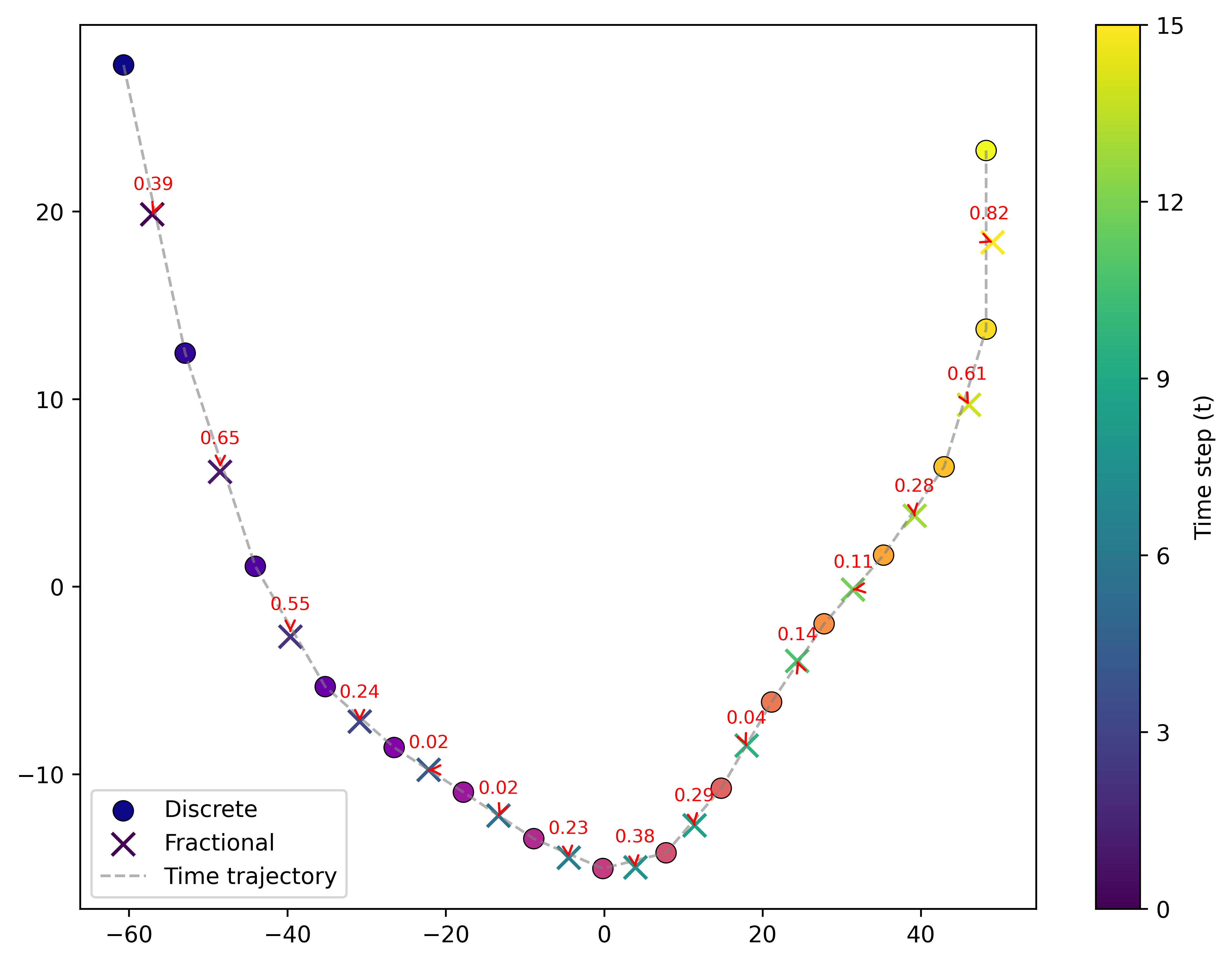}
\caption{PCA of \textit{TE} per discrete ($\bullet$) and fractional (\textit{X}) $\Delta t$. The number indicates deviation of a fractional from the interpolated value between the discrete \textit{TE}s.} 
\label{fig:two_col_images}
\end{figure}

Temporal sensitivity and sampling diversity of the learned prior are visualized in Figure~\ref{fig1}. In the left, we sampled 20 tokens using the prior for each $\Delta t$. The resulting trajectories align correctly with increasing time intervals, demonstrating that both the temporal encoder (TE) and the prior capture time sensitivity. Moreover, the principal component analysis (PCA) projection demonstrates that the prior distribution did not collapse to a single point, and was able to sample diverse and meaningful representations. This shows that even though the temporal encoding is deterministic with respect to the requested time interval, its effect on the underlying representation is not patient-independent. TE is added to the patient-specific image representation and then propagated through transformer self-attention and cross-attention. Therefore, the same $\Delta t$ can modulate different patients differently depending on their anatomical features and disease state.

To complement Figure~\ref{fig1}, we visualized the change in attention scores with different \textit{TE} (Fig.~\ref{fig:attmap}). The images indicate the highest activating regions w.r.t $CLS_t$. Since the input is a 3D volume with 3D ViT patchification, the attention maps are volumetric. We therefore visualize the central slice of the 3D attention map.
It can be observed that if \textit{TE} is removed from the encoder, the attention scores are more diffused around the image (Fig.~\ref{fig:attmap} Left), while with \textit{TE} they are more focused (Fig.~\ref{fig:attmap} Middle \& Right) on the retina. Another observation is that with different \textit{TE} prompting, 3D locations of the highest attention scores changed. This is in line with the temporal sensitivity of the representations, such that the visualization provides qualitative evidence that changing $\Delta t$ changes the spatial regions attended to by the model.

Also we evaluated the representation space under irregular time points by sampling with discrete (monthly), and fractional (every 15 days) time steps. Even though the model has never been trained with fractional time conditioning, 
in Figure~\ref{fig1} Right, it can be seen that the sampled tokens conditioned on fractional TE, follow the same time sensitive distribution. 
In Figure~\ref{fig:two_col_images}, TE values themselves are visualized with PCA. The fractional TEs deviate only slightly from the linearly interpolated values (in red).

\setlength{\tabcolsep}{3pt}
\begin{table}[!tbh]
\centering
\caption{Number of  parameters, FLOP per iteration and runtime per epoch of pretraining}\label{Tbl:flipflop}
{\fontsize{8}{10}\selectfont

\begin{tabular}{lccc}
\toprule
\textbf{Model} & \textbf{GFLOP} & \textbf{\# of Params (M)} & \textbf{Runtime (min)}\\ 
\midrule
MAE\cite{he2022masked}         &$45.1$ &$104.8$&$1.9$\\
SiamMAE\cite{gupta2023siamese}   &$153.6$ &$109.6$&$10.9$\\
CropMAE\cite{eymael2025efficient}  &$153.6$ &$109.6$&$10.9$\\
RSP\cite{jang2024visual}         &$281.9$ & $116.1$&$19.1$\\ 
STAMP        &$153.7$ &$117.4$& $10.6$\\
\bottomrule
\end{tabular}}
\end{table}

Finally, computational overhead for each method is presented in Table~\ref{Tbl:flipflop}. STAMP has insignificant computational overhead compared to SiamMAE. While RSP is more computationally expensive with 3 forward passes from its encoder, over long time windows of medical datasets, its performance drops considerably. On the other hand STAMP with simplified pipeline and time difference conditioning offers to pretrain the same ViT backbone better in a cost efficient manner.

\subsection{Risk Model}
STAMP pretraining allows the utilization of TE in a late-AMD conversion-risk model. With this, $\Delta t$ can be integrated as a covariate into the relative hazard function~\cite{kvamme2019time} (Equation~\ref{eq:risk}), while other backbones require specific architectural adaptations in the risk prediction head. As a result, STAMP outperformed the other baselines by a large margin (Table~\ref{Tbl:sv}).

\setlength{\tabcolsep}{3pt}
\begin{table*}[!th]
\centering
\caption{Linear evaluation of AD conversion prediction on pretrained models within two time windows: 1 and 3 years on the ADNI Dataset. The results are reported over repeated experiments}\label{Tbl:ADNI_Results}
{\fontsize{8}{10}\selectfont
\begin{tabular}{lccccccc}
\toprule
\multirow{2}{*}{\textbf{Model}} & \multicolumn{3}{c}{\textbf{1-year}} & \multicolumn{3}{c}{\textbf{3-years}} \\
\cmidrule(lr){2-4} \cmidrule(lr){5-7}
& AUROC $\uparrow$ & PRAUC $\uparrow$ & BACC $\uparrow$ & AUROC $\uparrow$ & PRAUC $\uparrow$ & BACC $\uparrow$ \\ \midrule
Random init  & $0.690\pm0.006$ & $0.142\pm0.002$ & $0.628\pm0.010$   & $0.713\pm0.030$ &$ 0.317\pm0.039$ & $0.652\pm0.015$   \\
ImageNet          & $0.736\pm0.007$ & $0.188\pm0.013$ &$0.639\pm0.008$ &$0.712\pm0.041$   & $0.301\pm0.056$ & $0.670\pm0.024$  \\
yAware\cite{dufumier2021contrastive}    & $0.730\pm0.001$ &$ 0.188\pm0.001$ & $0.625\pm0.009$  &   $0.716\pm0.004$ &$ 0.306\pm0.004$ & $0.629\pm0.004$    \\
MAE\cite{he2022masked}         & $0.721\pm0.002$ & $0.152\pm0.002$ & $0.634\pm0.012$& $0.712\pm0.002 $& $0.285\pm0.004$ & $0.674\pm0.011$ \\
SiamMAE\cite{gupta2023siamese}       & $0.775\pm0.004$    & $\underline{0.220\pm0.001}$&$0.721\pm0.006$  & $\mathbf{0.787\pm0.001}$    & $0.362\pm0.003$ & $\mathbf{0.726\pm0.002}$  \\ 
CropMAE\cite{eymael2025efficient}         & $0.769\pm0.004 $ & $0.198\pm0.001 $ & $0.712\pm0.002 $& $0.764\pm0.001 $& $\underline{0.394\pm0.001 }$ & $0.683\pm0.002 $ \\
RSP\cite{jang2024visual}        & \underline{$0.785\pm0.002 $}    & $0.215\pm0.002 $& \underline{$0.723\pm0.009 $}  &$0.738\pm0.002 $ & $0.339\pm0.004 $& $0.671\pm0.004 $    \\  \hline
STAMP           & $\mathbf{0.812\pm0.001}$ & $\mathbf{0.277\pm0.004}$ & $\mathbf{0.736\pm0.009}$ & $\underline{0.773\pm0.001}$ &$\mathbf{0.416\pm0.013}$ & $\underline{0.691\pm0.001}$  \\
\bottomrule
\end{tabular}}
\end{table*}

\setlength{\tabcolsep}{3pt}
\begin{table}[!th]
\centering
\caption{Linear evaluation of wet-AMD risk model. The models are pretrained with the HARBOR dataset. The results are reported in Concordance Index with bootstrapping}\label{Tbl:sv}
{\fontsize{8}{10}\selectfont

\begin{tabular}{lc}
\toprule
\textbf{Model} & \textbf{c-index $\uparrow$} \\ \midrule
TC\cite{emre2024learning}      &$\underline{0.664\pm0.023}$\\
MAE\cite{he2022masked}         & $0.604\pm0.074$\\
SiamMAE\cite{gupta2023siamese}   & $0.649\pm0.053$\\
CropMAE\cite{eymael2025efficient}        & $0.603\pm0.061$  \\
RSP\cite{jang2024visual}         & $0.599\pm0.146$ \\ \hline  
STAMP        & $\mathbf{0.747\pm0.071}$\\
\bottomrule
\end{tabular}}
\end{table}

\subsection{AD Progression}

We applied STAMP to 3D brain MRI using the ADNI dataset with irregular inter-visit time intervals across more than 60 centers (Table~\ref{Tbl:stat}), unlike HARBOR. Similar to the AMD progression prediction setup, the extracted features were linearly evaluated for AD progression prediction from MCI within 1 and 3 years (Table~\ref{Tbl:ADNI_Results}). Compared to AMD progression, ADNI evaluations highlight the generalizability of STAMP to different modalities, irregular visit schedules, and longer time windows. STAMP clearly outperformed other MAE based methods and a metadata-aware contrastive method~\cite{dufumier2021contrastive} in 1 year task across all three metrics. It achieved competitive results for the 3-year prediction task in terms of AUROC, while achieving the highest PRAUC, which is particularly important metric in the setting of class imbalance. Moreover, the lower performance of yAware~\cite{dufumier2021contrastive} relative to temporal MAE~\cite{gupta2023siamese,eymael2025efficient,jang2024visual}, further underscores the representation quality of MAE over contrastive approaches.

\section{Discussion \& Conclusion}
 We introduce STAMP, a self-supervised framework that learns temporal dynamics through time-conditioned stochasticity, enabling accurate disease progression prediction across different modalities and prognostic horizons with limited labels. By extending Siamese MAE pretraining with a time difference conditioned learnable prior and a future-aware posterior, STAMP effectively formulates forecasting as conditional variational inference to address the challenging aspects of disease progression prediction: long visit intervals, varying progression speeds, and multiple potential outcomes. The stochasticity and the time conditioning make it possible to prompt the model for possible futures at inference time, using a single scan and a specified time window. 

STAMP is better understood as a time-conditioned latent-dynamics pretraining objective for longitudinal 3D medical imaging. In latent-dynamics learning, a model learns to predict or sample a future latent state from the available current context, while future observations provide the training signal. STAMP adapts this principle to longitudinal medical imaging by using the time interval ($\Delta t$) as the transition condition and defining a time-conditioned prior ($p_\psi(z_{t+\Delta t}\mid h_t,\Delta t)$). This prior is directly available at inference, where only the current scan and the requested prediction horizon are known, enabling time-prompted stochastic representations for single-scan prognosis. Importantly, the stochastic latent is itself conditioned on $\Delta t$: the sampled disease state, not merely a deterministic embedding, depends on the requested interval, so different horizons yield different plausible future states. This dependency between the latent z and $\Delta t$ is what makes time-prompted sampling meaningful at inference.
 
We pretrained STAMP on two longitudinal datasets: 3D OCT scans for AMD and MRI scans from the ADNI cohort for Alzheimer’s disease onset. We evaluated pretrained models across multiple prognostic windows using the pretrained temporal encoding and stochastic sampling. After pretraining STAMP on HARBOR, STAMP consistently outperformed existing baselines and foundation models on 6 and 12 months progression for wet-AMD (HARBOR) and 12 months GA progression (PINNACLE) as a domain shift experiment. Similarly, on ADNI MRI (1 and 3 years AD onset), STAMP surpassed both MAE based and age-aware pretraining models, demonstrating robustness under irregular visit intervals and extended prediction windows.

The two-visit formulation is a deliberate design choice that keeps pretraining computationally scalable for 3D volumes. It also remains applicable in settings with as few as two visits per patient and aligns with the single-scan inference scenario. Pretraining on more than two volumes could further enhance its temporal capability; however it demands substantial computational resources for processing longitudinal 3D volumes. Moreover, the downstream task performance can be increased by predicting the onset with multiple visits, it would undermine the strengths of our model which is predicting the progression from a single visit utilizing the learned temporal and stochastic components. Additionally, the prior’s large sampling space makes post hoc explainability analysis of the sampled tokens challenging. 

As a future work, we plan to incorporate more than two visits using a temporal roll-out mechanism to account for longer disease trajectories. This would enable step-wise sampling with different temporal intervals, allowing the model to learn richer stochastic disease dynamics over complete trajectories.

Clinically, STAMP helps with utilizing large unlabeled datasets to improve personalized monitoring by forecasting individual disease trajectories from a single baseline scan, facilitating tailored follow-up schedules, earlier intervention for high-risk patients, and reducing hospital load. This demonstrates the promise of self-supervised, time-conditioned stochastic modeling for advancing precision medicine in progressive diseases.

\section*{Acknowledgment}

Data collection and sharing for this project was funded by the Alzheimer's Disease Neuroimaging Initiative (ADNI) (National Institutes of Health Grant U01 AG024904) and DOD ADNI (Department of Defense award number W81XWH-12-2-0012). ADNI is funded by the National Institute on Aging, the National Institute of Biomedical Imaging and Bioengineering, and through generous contributions from the following: AbbVie, Alzheimer’s Association; Alzheimer’s Drug Discovery Foundation; Araclon Biotech; BioClinica, Inc.; Biogen; Bristol-Myers Squibb Company; CereSpir, Inc.; Cogstate; Eisai Inc.; Elan Pharmaceuticals, Inc.; Eli Lilly and Company; EuroImmun; F. Hoffmann-La Roche Ltd and its affiliated company Genentech, Inc.; Fujirebio; GE Healthcare; IXICO Ltd.; Janssen Alzheimer Immunotherapy Research \& Development, LLC.; Johnson \& Johnson Pharmaceutical Research \& Development LLC.; Lumosity; Lundbeck; Merck \& Co., Inc.; Meso Scale Diagnostics, LLC.; NeuroRx Research; Neurotrack Technologies; Novartis Pharmaceuticals Corporation; Pfizer Inc.; Piramal Imaging; Servier; Takeda Pharmaceutical Company; and Transition Therapeutics. The Canadian Institutes of Health Research is providing funds to support ADNI clinical sites in Canada. Private sector contributions are facilitated by the Foundation for the National Institutes of Health (www.fnih.org). The grantee organization is the Northern California Institute for Research and Education, and the study is coordinated by the Alzheimer’s Therapeutic Research Institute at the University of Southern California. ADNI data are disseminated by the Laboratory for Neuro Imaging at the University of Southern California.

\bibliographystyle{IEEE}
\bibliography{bibs}

@inproceedings{huo2025time,
  title={Time-to-event pretraining for 3d medical imaging},
  author={Huo, Zepeng Frazier and Fries, Jason and Lozano, Alejandro and Valanarasu, Jeya Maria Jose and Steinberg, Ethan and Blankemeier, Louis and Chaudhari, Akshay and Langlotz, Curtis and Shah, Nigam},
  booktitle={International Conference on Learning Representations},
  volume={2025},
  pages={84072--84108},
  year={2025}
}

@article{holste2024harnessing,
  title={Harnessing the power of longitudinal medical imaging for eye disease prognosis using Transformer-based sequence modeling},
  author={Holste, Gregory and Lin, Mingquan and Zhou, Ruiwen and Wang, Fei and Liu, Lei and Yan, Qi and Van Tassel, Sarah H and Kovacs, Kyle and Chew, Emily Y and Lu, Zhiyong and others},
  journal={NPJ Digital Medicine},
  volume={7},
  number={1},
  pages={216},
  year={2024},
  publisher={Nature Publishing Group UK London}
}

@article{kvamme2019time,
  title={Time-to-event prediction with neural networks and Cox regression},
  author={Kvamme, H{\aa}vard and Borgan, {\O}rnulf and Scheel, Ida},
  journal={Journal of machine learning research},
  volume={20},
  number={129},
  pages={1--30},
  year={2019}
}

@article{andersen1982cox,
  title={Cox's regression model for counting processes: a large sample study},
  author={Andersen, Per Kragh and Gill, Richard D},
  journal={The annals of statistics},
  pages={1100--1120},
  year={1982},
  publisher={JSTOR}
}

@article{zhang2018time,
  title={Time-varying covariates and coefficients in Cox regression models},
  author={Zhang, Zhongheng and Reinikainen, Jaakko and Adeleke, Kazeem Adedayo and Pieterse, Marcel E and Groothuis-Oudshoorn, Catharina GM},
  journal={Annals of translational medicine},
  volume={6},
  number={7},
  pages={121},
  year={2018}
}

@inproceedings{
micheli2024efficient,
title={Efficient World Models with Context-Aware Tokenization},
author={Vincent Micheli and Eloi Alonso and François Fleuret},
booktitle={International Conference on Machine Learning},
pages={35623 -- 35638},
year={2024},
  organization={PMLR}
}

@inproceedings{
loshchilov2018decoupled,
title={Decoupled Weight Decay Regularization},
author={Ilya Loshchilov and Frank Hutter},
booktitle={International Conference on Learning Representations},
year={2019},
}

@article{sutton2023developing,
  title={Developing and validating a multivariable prediction model which predicts progression of intermediate to late age-related macular degeneration—the PINNACLE trial protocol},
  author={Sutton, Janice and others},
  journal={Eye},
  volume={37},
  number={6},
  pages={1275--1283},
  year={2023},
  publisher={Nature Publishing Group UK London}
}

@article{kingma2014semi,
  title={Semi-supervised learning with deep generative models},
  author={Kingma, Diederik P and Rezende, Danilo J and Mohamed, Shakir and Welling, Max},
  journal={Advances in neural information processing systems},
  volume={27},
pages = {3581–3589},
  year={2014}
}

@misc{kingma2013auto,
  title={Auto-encoding variational bayes},
  author={Kingma, Diederik P and Welling, Max},
  year={2013},
  publisher={Banff, Canada}
}

@article{huang2023self,
  title={Self-supervised learning for medical image classification: a systematic review and implementation guidelines},
  author={Huang, Shih-Cheng and Pareek, Anuj and Jensen, Malte and Lungren, Matthew P and Yeung, Serena and Chaudhari, Akshay S},
  journal={NPJ Digital Medicine},
  volume={6},
  number={1},
  pages={74},
  year={2023},
  publisher={Nature Publishing Group UK London}
}

@article{ren2022local,
  title={Local spatiotemporal representation learning for longitudinally-consistent neuroimage analysis},
  author={Ren, Mengwei and Dey, Neel and Styner, Martin and Botteron, Kelly and Gerig, Guido},
  journal={Advances in Neural Information Processing Systems},
  volume={35},
  pages={13541--13556},
  year={2022}
}

@article{bucholc2023hybrid,
  title={A hybrid machine learning approach for prediction of conversion from mild cognitive impairment to dementia},
  author={Bucholc, Magda and Titarenko, Sofya and Ding, Xuemei and Canavan, Callum and Chen, Tianhua},
  journal={Expert Systems with Applications},
  volume={217},
  pages={119541},
  year={2023},
  publisher={Elsevier}
}

@article{breijyeh2020comprehensive,
  title={Comprehensive review on Alzheimer’s disease: causes and treatment},
  author={Breijyeh, Zeinab and Karaman, Rafik},
  journal={Molecules},
  volume={25},
  number={24},
  pages={5789},
  year={2020},
  publisher={MDPI}
}

@article{BERG2015146,
title = {Comparison of Ranibizumab and Bevacizumab for Neovascular Age-Related Macular Degeneration According to LUCAS Treat-and-Extend Protocol},
journal = {Ophthalmology},
volume = {122},
number = {1},
pages = {146-152},
year = {2015},
issn = {0161-6420},
doi = {https://doi.org/10.1016/j.ophtha.2014.07.041},
author = {Karina Berg and Terje R. Pedersen and Leiv Sandvik and Ragnheiður Bragadóttir}
}

@article{hallak2019imaging,
  title={Imaging, genetic, and demographic factors associated with conversion to neovascular age-related macular degeneration: secondary analysis of a randomized clinical trial},
  author={Hallak, Joelle A and de Sisternes, Luis and Osborne, Aaron and Yaspan, Brian and Rubin, Daniel L and Leng, Theodore},
  journal={JAMA ophthalmology},
  volume={137},
  number={7},
  pages={738--744},
  year={2019},
  publisher={American Medical Association}
}

@inproceedings{chen2023masked,
  title={Masked image modeling advances 3d medical image analysis},
  author={Chen, Zekai and Agarwal, Devansh and Aggarwal, Kshitij and Safta, Wiem and Balan, Mariann Micsinai and Brown, Kevin},
  booktitle={Proceedings of the IEEE/CVF Winter Conference on Applications of Computer Vision},
  pages={1970--1980},
  year={2023}
}

@inproceedings{
burchi2025accurate,
title={Accurate and Efficient World Modeling with Masked Latent Transformers},
author={Maxime Burchi and Radu Timofte},
booktitle={International Conference on Machine Learning},
organization={PMLR},
year={2025},
}

@inproceedings{wang2024dust3r,
  title={Dust3r: Geometric 3d vision made easy},
  author={Wang, Shuzhe and Leroy, Vincent and Cabon, Yohann and Chidlovskii, Boris and Revaud, Jerome},
  booktitle={Proceedings of the IEEE/CVF Conference on Computer Vision and Pattern Recognition},
  pages={20697--20709},
  year={2024}
}

@article{petersen2010alzheimer,
  title={Alzheimer's disease Neuroimaging Initiative (ADNI) clinical characterization},
  author={Petersen, Ronald Carl and others},
  journal={Neurology},
  volume={74},
  number={3},
  pages={201--209},
  year={2010},
  publisher={Lippincott Williams \& Wilkins}
}

@inproceedings{dufumier2021contrastive,
  title={Contrastive learning with continuous proxy meta-data for 3d mri classification},
  author={Dufumier, Benoit and others},
  booktitle={International Conference on Medical Image Computing and Computer-Assisted Intervention},
  pages={58--68},
  year={2021},
  organization={Springer}
}

@inproceedings{carreira2017quo,
  title={Quo vadis, action recognition? a new model and the kinetics dataset},
  author={Carreira, Joao and Zisserman, Andrew},
  booktitle={Proceedings of the IEEE/CVF Conference on Computer Vision and Pattern Recognition},
  pages={6299--6308},
  year={2017}
}

@article{avants2008symmetric,
  title={Symmetric diffeomorphic image registration with cross-correlation: evaluating automated labeling of elderly and neurodegenerative brain},
  author={Avants, Brian B and Epstein, Charles L and Grossman, Murray and Gee, James C},
  journal={Medical image analysis},
  volume={12},
  number={1},
  pages={26--41},
  year={2008},
  publisher={Elsevier}
}

@article{hoopes2022synthstrip,
  title={SynthStrip: skull-stripping for any brain image},
  author={Hoopes, Andrew and Mora, Jocelyn S and Dalca, Adrian V and Fischl, Bruce and Hoffmann, Malte},
  journal={NeuroImage},
  volume={260},
  pages={119474},
  year={2022},
  publisher={Elsevier}
}

@article{tustison2010n4itk,
  title={N4ITK: improved N3 bias correction},
  author={Tustison, Nicholas J and others},
  journal={IEEE transactions on medical imaging},
  volume={29},
  number={6},
  pages={1310--1320},
  year={2010},
  publisher={IEEE}
}

@inproceedings{kunanbayev2024training,
  title={Training ViT with Limited Data for Alzheimer’s Disease Classification: An Empirical Study},
  author={Kunanbayev, Kassymzhomart and Shen, Vyacheslav and Kim, Dae-Shik},
  booktitle={International Conference on Medical Image Computing and Computer-Assisted Intervention},
  pages={334--343},
  year={2024},
  organization={Springer}
}

@article{ansart2021predicting,
  title={Predicting the progression of mild cognitive impairment using machine learning: A systematic, quantitative and critical review},
  author={Ansart, Manon and others},
  journal={Medical Image Analysis},
  volume={67},
  pages={101848},
  year={2021},
  publisher={Elsevier}
}

@inproceedings{puglisi2024enhancing,
  title={Enhancing spatiotemporal disease progression models via latent diffusion and prior knowledge},
  author={Puglisi, Lemuel and Alexander, Daniel C and Rav{\`\i}, Daniele},
  booktitle={International Conference on Medical Image Computing and Computer-Assisted Intervention},
  pages={173--183},
  year={2024},
  organization={Springer}
}

@article{elfwing2018sigmoid,
  title={Sigmoid-weighted linear units for neural network function approximation in reinforcement learning},
  author={Elfwing, Stefan and Uchibe, Eiji and Doya, Kenji},
  journal={Neural networks},
  volume={107},
  pages={3--11},
  year={2018},
  publisher={Elsevier}
}

@inproceedings{bertasius2021space,
    author  = {Gedas Bertasius and Heng Wang and Lorenzo Torresani},
    title = {Is Space-Time Attention All You Need for Video Understanding?},
    booktitle={International Conference on Machine Learning},
    organization={PMLR},
    pages = {813--824},
    month = {July},
    year = {2021}
}

@article{cong2022satmae,
  title={Satmae: Pre-training transformers for temporal and multi-spectral satellite imagery},
  author={Cong, Yezhen and others},
  journal={Advances in Neural Information Processing Systems},
  volume={35},
  pages={197--211},
  year={2022}
}

@inproceedings{
schmidt2024learning,
title={Learning to Act without Actions},
author={Dominik Schmidt and Minqi Jiang},
booktitle={The Twelfth International Conference on Learning Representations},
year={2024},
}

@article{bengio2013estimating,
  title={Estimating or propagating gradients through stochastic neurons for conditional computation},
  author={Bengio, Yoshua and L{\'e}onard, Nicholas and Courville, Aaron},
  journal={arXiv preprint arXiv:1308.3432},
  year={2013}
}

@article{visionfm,
author = {Jianing Qiu  and others },
title = {Development and Validation of a Multimodal Multitask Vision Foundation Model for Generalist Ophthalmic Artificial Intelligence},
journal = {NEJM AI},
volume = {1},
number = {12},
pages = {AIoa2300221},
year = {2024},
}

@inproceedings{elscalable,
  title     = {Scalable Pre-training of Large Autoregressive Image Models},
  author    = {El-Nouby, Alaaeldin and others},
    booktitle={International Conference on Machine Learning},
pages = 	 {12371--12384},
    organization={PMLR},
  year      = {2024},
}

@article{chen2024context,
  title={Context autoencoder for self-supervised representation learning},
  author={Chen, Xiaokang and others},
  journal={International Journal of Computer Vision},
  volume={132},
  number={1},
  pages={208--223},
  year={2024},
  publisher={Springer}
}

@inproceedings{emre2024learning,
  title={Learning Temporally Equivariance for Degenerative Disease Progression in OCT by Predicting Future Representations},
  author={Emre, Taha and Chakravarty, Arunava and Lachinov, Dmitrii and Rivail, Antoine and Schmidt-Erfurth, Ursula and Bogunovi{\'c}, Hrvoje},
  booktitle={International Conference on Medical Image Computing and Computer-Assisted Intervention},
  pages={196--206},
  year={2024},
  organization={Springer}
}

@inproceedings{peebles2023scalable,
  title={Scalable diffusion models with transformers},
  author={Peebles, William and Xie, Saining},
  booktitle={Proceedings of the IEEE/CVF International Conference on Computer Vision},
  pages={4195--4205},
  year={2023}
}

@inproceedings{
jang2024visual,
title={Visual Representation Learning with Stochastic Frame Prediction},
author={Huiwon Jang and Dongyoung Kim and Junsu Kim and Jinwoo Shin and Pieter Abbeel and Younggyo Seo},
booktitle={International Conference on Machine Learning},
pages = 	 {21289--21305},
organization={PMLR},
year={2024},
}

@inproceedings{petrovich2021action,
  title={Action-conditioned 3d human motion synthesis with transformer vae},
  author={Petrovich, Mathis and Black, Michael J and Varol, G{\"u}l},
  booktitle={Proceedings of the IEEE/CVF International Conference on Computer Vision},
  pages={10985--10995},
  year={2021}
}

@inproceedings{denton2018stochastic,
  title={Stochastic video generation with a learned prior},
  author={Denton, Emily and Fergus, Rob},
  booktitle={International conference on machine learning},
  pages={1174--1183},
  year={2018},
  organization={PMLR}
}

@article{YANG20191,
title = {Two-Year Risk of Exudation in Eyes with Nonexudative Age-Related Macular Degeneration and Subclinical Neovascularization Detected with Swept Source Optical Coherence Tomography Angiography},
journal = {American Journal of Ophthalmology},
volume = {208},
pages = {1-11},
year = {2019},
issn = {0002-9394},
author = {Jin Yang and others},
}

@inproceedings{eymael2025efficient,
  title={Efficient image pre-training with siamese cropped masked autoencoders},
  author={Eyma{\"e}l, Alexandre and Vandeghen, Renaud and Cioppa, Anthony and Giancola, Silvio and Ghanem, Bernard and Van Droogenbroeck, Marc},
  booktitle={European Conference on Computer Vision},
  pages={348--366},
  year={2025},
  organization={Springer}
}

@article{zeghlache2025mae,
  title={L-MAE: Longitudinal masked auto-encoder with time and severity-aware encoding for diabetic retinopathy progression prediction},
  author={Zeghlache, Rachid and others},
  journal={Computers in Biology and Medicine},
  volume={185},
  pages={109508},
  year={2025},
  publisher={Elsevier}
}

@inproceedings{zhang2024whole,
  title={Whole heart 3d+ t representation learning through sparse 2d cardiac mr images},
  author={Zhang, Yundi and Chen, Chen and Shit, Suprosanna and Starck, Sophie and Rueckert, Daniel and Pan, Jiazhen},
  booktitle={International Conference on Medical Image Computing and Computer-Assisted Intervention},
  pages={359--369},
  year={2024},
  organization={Springer}
}

@inproceedings{kong2023understanding,
  title={Understanding masked autoencoders via hierarchical latent variable models},
  author={Kong, Lingjing and others},
  booktitle={Proceedings of the IEEE/CVF Conference on Computer Vision and Pattern Recognition},
  pages={7918--7928},
  year={2023}
}

@article{tang2023crossscale,
  title={Cross-scale mae: A tale of multiscale exploitation in remote sensing},
  author={Tang, Maofeng and Cozma, Andrei and Georgiou, Konstantinos and Qi, Hairong},
  journal={Advances in Neural Information Processing Systems},
  volume={36},
  pages={20054--20066},
  year={2023}
}

@article{weinzaepfel2022croco,
  title={Croco: Self-supervised pre-training for 3d vision tasks by cross-view completion},
  author={Weinzaepfel, Philippe and others},
  journal={Advances in Neural Information Processing Systems},
  volume={35},
  pages={3502--3516},
  year={2022}
}

@article{hafner2023mastering,
  title={Mastering diverse domains through world models},
  author={Hafner, Danijar and Pasukonis, Jurgis and Ba, Jimmy and Lillicrap, Timothy},
  journal={arXiv preprint arXiv:2301.04104},
  year={2023}
}

@inproceedings{
dosovitskiy2021an,
title={An Image is Worth 16x16 Words: Transformers for Image Recognition at Scale},
author={Alexey Dosovitskiy and others},
booktitle={International Conference on Learning Representations},
year={2021},
}

@inproceedings{wu2023dropmae,
  title={Dropmae: Masked autoencoders with spatial-attention dropout for tracking tasks},
  author={Wu, Qiangqiang and Yang, Tianyu and Liu, Ziquan and Wu, Baoyuan and Shan, Ying and Chan, Antoni B},
  booktitle={Proceedings of the IEEE/CVF Conference on Computer Vision and Pattern Recognition},
  pages={14561--14571},
  year={2023}
}

@article{feichtenhofer2022masked,
  title={Masked autoencoders as spatiotemporal learners},
  author={Feichtenhofer, Christoph and Li, Yanghao and He, Kaiming},
  journal={Advances in neural information processing systems},
  volume={35},
  pages={35946--35958},
  year={2022}
}

@inproceedings{pathak2016context,
  title={Context encoders: Feature learning by inpainting},
  author={Pathak, Deepak and Krahenbuhl, Philipp and Donahue, Jeff and Darrell, Trevor and Efros, Alexei A},
  booktitle={Proceedings of the IEEE conference on computer vision and pattern recognition},
  pages={2536--2544},
  year={2016}
}

@inproceedings{
tong2022videomae,
title={Video{MAE}: Masked Autoencoders are Data-Efficient Learners for Self-Supervised Video Pre-Training},
author={Zhan Tong and Yibing Song and Jue Wang and Limin Wang},
booktitle={Advances in Neural Information Processing Systems},
year={2022},
volume={35},
  pages={10078--10093},
}

@article{gupta2023siamese,
  title={Siamese masked autoencoders},
  author={Gupta, Agrim and Wu, Jiajun and Deng, Jia and Li, Fei-Fei},
  journal={Advances in Neural Information Processing Systems},
  volume={36},
  pages={40676--40693},
  year={2023}
}

@article{zhou2023retfound,
  title={A foundation model for generalizable disease detection from retinal images},
  author={Zhou, Yukun and others},
  journal={Nature},
  volume={622},
  number={7981},
  pages={156--163},
  year={2023},
  publisher={Nature Publishing Group UK London}
}

@inproceedings{he2022masked,
  title={Masked autoencoders are scalable vision learners},
  author={He, Kaiming and Chen, Xinlei and Xie, Saining and Li, Yanghao and Doll{\'a}r, Piotr and Girshick, Ross},
  booktitle={Proceedings of the IEEE/CVF conference on computer vision and pattern recognition},
  pages={16000--16009},
  year={2022}
}

@inproceedings{chen2020simple,
  title={A simple framework for contrastive learning of visual representations},
  author={Chen, Ting and Kornblith, Simon and Norouzi, Mohammad and Hinton, Geoffrey},
  booktitle={International Conference on Machine Learning},
  pages={1597--1607},
  year={2020},
  organization={PMLR}
}

@inproceedings{rivail2019modeling,
  title={Modeling disease progression in retinal OCTs with longitudinal self-supervised learning},
  author={Rivail, Antoine and others},
  booktitle={International Workshop on PRedictive Intelligence In MEdicine},
  pages={44--52},
  year={2019},
  organization={Springer}
}

@article{yim2020predicting,
  title={Predicting conversion to wet age-related macular degeneration using deep learning},
  author={Yim, Jason and others},
  journal={Nature Medicine},
  volume={26},
  number={6},
  pages={892--899},
  year={2020},
  publisher={Nature Publishing Group}
}

@article{bressAMD,
    author = {Bressler, Neil M.},
    title = "{Age-Related Macular Degeneration Is the Leading Cause of Blindness}",
    journal = {JAMA},
    volume = {291},
    number = {15},
    pages = {1900-1901},
    year = {2004},
    month = {04},
    issn = {0098-7484},
}

\end{document}